\documentclass[runningheads]{llncs}

\usepackage[T1]{fontenc}
\usepackage{graphicx}
\usepackage{hyperref}
\usepackage{subcaption}
\usepackage[font=small,labelfont=bf]{caption}
\usepackage{xcolor}
\usepackage{tikz}
\usepackage{amsmath}
\usepackage[ruled,vlined,linesnumbered]{algorithm2e}
\usepackage{pgffor}
\usepackage{siunitx}
\usetikzlibrary{shapes.geometric, arrows.meta, positioning, calc, fit, backgrounds}
\newlength{\envColWidth}

\newcommand{\cbox}[1]{\textcolor{#1}{\rule{0.8em}{0.8em}}}
\newcommand{\legendline}[1]{\textcolor{#1}{\rule{1.2em}{1.2pt}}}
\newcommand{\cmark}[1]{\textcolor{#1}{\rule{0.6em}{0.6em}}}
\newcommand{\csquare}[1]{\textcolor{#1}{\rule{0.8em}{0.8em}}}

\def\V{V}
\SetKwData{failure}{failure}
\def\xstart{x_{\text{start}}}
\def\Xgoal{X_{\text{goal}}}

\def\parameterFiboJitter{\ensuremath{p_{\text{fibo-jitter}}}}
\def\parameterFiboJitterValue{\ensuremath{\frac{\pi}{8}}}

\def\parameterUniformValidReward{\ensuremath{c_u}}
\def\parameterUniformValidRewardValue{\ensuremath{\num{1e8}}}

\def\parameterPCAValidReward{\ensuremath{c_s}}
\def\parameterPCAValidRewardValue{\ensuremath{\num{5.0}}}

\def\parameterMABSlidingWindow{\ensuremath{N_{\text{sliding-window}}}}
\def\parameterMABSlidingWindowValue{\ensuremath{256}}

\def\parameterScaleSamplingBatchSize{\ensuremath{b}}
\def\parameterScaleSamplingBatchSizeValue{\ensuremath{64}}

\def\parameterScaleSamplingInitialRadius{\ensuremath{r_0}}
\def\parameterScaleSamplingInitialRadiusValue{\ensuremath{1.0}}

\def\parameterScaleSamplingMinRadius{\ensuremath{r_{\text{min}}}}
\def\parameterScaleSamplingMinRadiusValue{\ensuremath{\num{1e-6}}}

\def\parameterScaleSamplingMaxRadius{\ensuremath{r_{\text{max}}}}
\def\parameterScaleSamplingMaxRadiusValue{\ensuremath{25.0}}

\def\parameterScaleSamplingGrowthFactor{\ensuremath{g}}
\def\parameterScaleSamplingGrowthFactorValue{\ensuremath{\exp{{(-0.7)}}}}

\def\parameterScaleSamplingShrinkFactor{\ensuremath{s}}
\def\parameterScaleSamplingShrinkFactorValue{\ensuremath{\exp{{(0.9)}}}}

\def\parameterScaleSamplingMaxSteps{\ensuremath{S}}
\def\parameterScaleSamplingMaxStepsValue{\ensuremath{50}}

\def\parameterScaleSamplingMinValidityRate{\ensuremath{\alpha_{\text{min}}}}
\def\parameterScaleSamplingMinValidityRateValue{\ensuremath{0.1}}

\def\parameterScaleSamplingMaxValidityRate{\ensuremath{\alpha_{\text{max}}}}
\def\parameterScaleSamplingMaxValidityRateValue{\ensuremath{0.5}}

\definecolor{mabrrtcolor}{HTML}{EE0000}
\definecolor{bridgerrtcolor}{HTML}{0000FF}
\definecolor{gaussianrrtcolor}{HTML}{9932CC}
\definecolor{obstaclerrtcolor}{HTML}{228B22}
\definecolor{matevectrrtcolor}{HTML}{FF8C00}
\definecolor{bkrrtcolor}{HTML}{8B4513}
\definecolor{bfscolor}{HTML}{00AEEF}

\title{Scale-Invariant Sampling for Sub-Millimeter Object Extraction from Narrow Passages}
\title{Motion Planning for Object Extraction using Scale-Invariant Sampling in Multi-Arm Bandits}
\title{Scale-Invariant Sampling in Multi-Arm Bandit Motion Planning for Object Extraction}
\titlerunning{Scale-Invariant Sampling}
\author{
Servet B. Bayraktar\inst{1} \and
Andreas Orthey\inst{1} \and
Marc Toussaint\inst{1}
}

\authorrunning{Bayraktar et al.}

\institute{
Technical University of Berlin
}

\begin{document}

\maketitle
% ---------- Content ----------
\begin{abstract}
Object extraction tasks often occur in disassembly problems, where bolts, screws, or pins have to be removed from tight, narrow spaces. 
In such problems, the distance to the environment is often on the millimeter scale. 
Sampling-based planners can solve such problems and provide completeness guarantees. 
However, sampling becomes a bottleneck, since almost all motions will result in collisions with the environment. 
To overcome this problem, we propose a novel scale-invariant sampling strategy which explores the configuration space using a grow-shrink search to find useful, high-entropy sampling scales. 
Once a useful sampling scale has been found, our framework exploits this scale by using a principal components analysis (PCA) to find useful directions for object extraction.
We embed this sampler into a multi-arm bandit rapidly-exploring random tree (MAB-RRT) planner and test it on eight challenging 3D object extraction scenarios, involving bolts, gears, rods, pins, and sockets. 
To evaluate our framework, we compare it with classical sampling strategies like uniform sampling, obstacle-based sampling, and narrow-passage sampling, and with modern strategies like mate vectors, physics-based planning, and disassembly breadth first search. 
Our experiments show that scale-invariant sampling improves success rate by one order of magnitude on 7 out of 8 scenarios. This demonstrates that scale-invariant sampling is an important concept for general purpose object extraction in disassembly tasks.
\end{abstract}
\section{Introduction}

Removing an object out of a tight, narrow passage is a fundamental skill for robot disassembly tasks in recycling~\cite{asif2024robotic}, repair~\cite{parker1998robotics}, and remanufacturing~\cite{laili2022optimisation,das2025towards}.
Sampling-based methods~\cite{Orthey2023AnnualReview} can tackle
such problems and provide completeness
guarantees~\cite{zickler2009efficient,aguinaga2008targetless,Ebinger2018MateVecTRRT,tian2022assemble}. 
However, due to narrow
passages, such problems often become intractable to solve~\cite{Ebinger2018MateVecTRRT}. This is often due to the large scale on which methods
like rapidly exploring random tree (RRT)~\cite{Kuffner2000} operate, where motions out of narrow passages are almost always invalid.

To tackle this problem, we propose a novel scale-invariant sampling strategy.
This strategy is based on the observation that scale is an important
consideration for motion planning of object extraction tasks. For example, if a point robot is inside a
narrow passage, and you sample points inside a large radius around it, then almost all of the points
will be unreachable (the robot just bumps into the wall). In contrast, if we sample in an infinitesimal small radius,
then almost all samples will be reachable. Those two scales have,
however, a low information entropy~\cite{jaynes2003probability} and do not help
a planner to make good decisions. What this implies, however, is that there is a
scale, at which the information entropy is high, meaning samples are maximally useful for planning. Such a scale corresponds to
situations where roughly half of the samples become reachable~\cite{jaynes2003probability}. This scale
would not only have a high information entropy, but it would be an exceptional guidance to the planner.

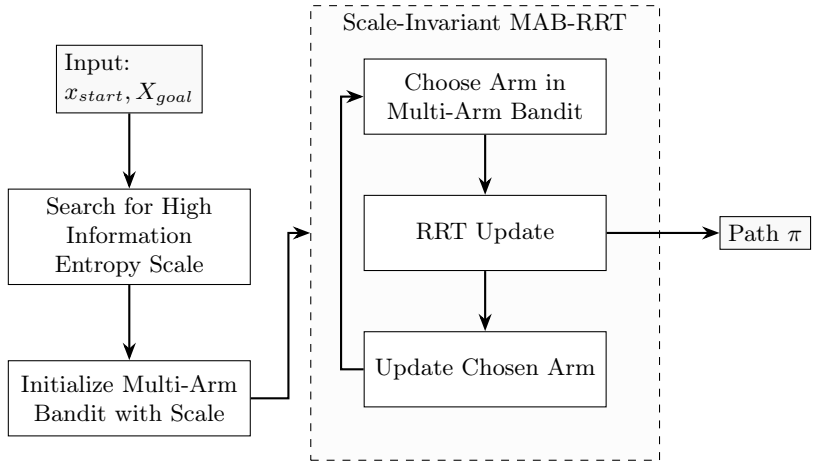
\begin{figure}
    \centering
\begin{tikzpicture}[
    node distance=1.0cm and 1.0cm,
    block/.style={rectangle, draw, fill=white, text width=3cm, align=center, minimum height=1cm, font=\small},
    io/.style={draw, fill=gray!5, font=\small},
    loop_box/.style={draw, dashed, inner sep=20pt, fill=gray!2},
    arrow/.style={-Stealth, thick}
]

    % --- Inputs ---
    \node (input) [io, align=left] {Input: \\ $x_{start}, X_{goal}$};

    % --- First Stage ---
    \node (entropy) [block, below=of input] {Search for High Information Entropy Scale};

    % --- Second Stage ---
    \node (bandit_init) [block, below=of entropy] {Initialize Multi-Arm Bandit with Scale};

    % --- Loop Container ---
    % Sub-blocks inside the loop
    \node (sample) [block, above right=3cm and 1.5cm of bandit_init] {Choose Arm in Multi-Arm Bandit};
    \node (rrt) [block, below=0.8cm of sample] {RRT Update};
    \node (update) [block, below=0.8cm of rrt] {Update Chosen Arm};

    % The loop box/container
    \begin{scope}[on background layer]
        \node (loop_container) [loop_box, fit=(sample) (rrt) (update), label={[anchor=north]north:Scale-Invariant MAB-RRT}] {};
    \end{scope}

    % --- Output ---
    \node (output) [io, right=1.5cm of rrt] {Path $\pi$};

    % --- Connections ---
    \draw [arrow] (input) -- (entropy);
    \draw [arrow] (entropy) -- (bandit_init);
    
    % Connection to the loop
    \draw [arrow] (bandit_init.east) -- ++(0.5,0) |- (loop_container.west);

    % Internal loop flow
    \draw [arrow] (sample) -- (rrt);
    \draw [arrow] (rrt) -- (update);
    
    % Feedback loop arrow
    \draw [arrow] (update.west) -- ++(-0.3,0) |- (sample.west);

    % Connection to output
    \draw [arrow] (rrt.east) -- (output.west);

\end{tikzpicture}
    \caption{Overview about scale-invariant sampling in MAB-RRT. As input, we use a start configuration $\xstart$ and a goal region $\Xgoal$. We then search for a high-information entropy scale and initialize the multi-arm bandit with this scale. We then start a MAB-RRT loop, where an arm is chose, the chosen sampler is used for one RRT update step, and the arm reward is updated. This loop continues until a termination condition is true or a path $\pi$ has been found which solves the problem.}
    \label{fig:system}
\end{figure}

To implement this insight, we devise two sampling strategies. One is a scale
sampler, which explores different scales to find a high information
entropy scale. The other is a principal component analysis (PCA) sampler, which exploits the principal
components in the samples at the previously computed high information entropy scale. Both of those
samplers are integrated into a multi-arm bandit RRT
(MAB-RRT)~\cite{faroni2023motion}. An overview about this framework is shown in Fig.~\ref{fig:system}. We found scale-invariant sampling in MAB-RRT to be an effective planning
strategy for disassembly tasks where bolts, pins, or gears have to be removed from a narrow passage. 

In summary, our contributions are:
\begin{enumerate}
    \item We propose a novel scale sampler which can autonomously find high
      information entropy scales for efficient sampling densities.
    \item We develop a PCA sampler to exploit a given high information entropy scale by biasing samples
      along a positive or negative principal component direction.
    \item Embedding both samplers into the MAB-RRT planner to combine it
      with other sampling strategies like uniform sampling.
    \item We provide an implementation of samplers and MAB-RRT as an extension
      of the Open Motion Planning Library (OMPL) and provide a set of demos scenarios.
\end{enumerate}

Eventually, this sampling strategy is tested against other classical sampling
methods on eight challenging 3D scenarios involving realistic object extraction
tasks.

\section{Related Work}
This work is closely related to object extraction, biased sampling in motion
planning, and PCA-based sampling. We
review those topics and discuss how our work is connected.

\subsection{Object Extraction}

Object extraction is the task of separating two objects until the distance
between them reaches a predefined threshold. Our interest is in tasks where the
objects have to move along tight, narrow passages to achieve separation, a
task which is often encountered in disassembly problems~\cite{tian2022assemble}.

A main focus of research in this area is on fastener removal. For example,
detecting screw types and using the correct tool for screw
extraction~\cite{zhang2023automatic} or learning policies for
single object extraction tasks with a robot
manipulator~\cite{serrano2023learning}. Specialized systems exists for specific
screw types like hexagonal screws~\cite{li2020unfastening}.
Our work is complementary in that we tackle arbitrary objects, including bolts,
nuts, or gears, for which a removal plan has first to be computed to achieve a
separation.

Apart from fastener removal, research has also focused on cluttered extraction. 
This includes extracting boxes from piles without other boxes
collapsing~\cite{pathak2025collapse}, or carefully extracting objects from a
pile without disturbing the other objects~\cite{Motoda2023}.

For object extraction in disassemblies, specialized planners have been
developed. For example, the Targetless-RRT~\cite{aguinaga2008targetless}
explores the space without a specified goal region, but instead tries to
minimize its distance to a predefined exterior space which represents a
disassembled state. This idea has later been combined with mating vectors in the
Mating Vector and Targetless-RRT
(MateVec-TRRT)~\cite{Ebinger2018MateVecTRRT,tian2022assemble}. MateVec-TRRT uses
the concept of mating vectors, which are vectors which point into the direction
of a separation between two objects. This is particularly advantageous when two
objects are close by and just have to be pulled
apart~\cite{Ebinger2018MateVecTRRT}. Another approach is to use physics-based
separation vectors~\cite{tian2022assemble,zickler2009efficient}. This idea is
based on the observation that nuts and bolts can be moved by an approximate
force which has a component into the motion direction~\cite{tian2022assemble}.
This avoids costly computation of the exact motion direction. This can then be
implemented in a planner like the Behavioral Kinodynamic Rapidly-Exploring
Random Trees (BK-RRT)~\cite{zickler2009efficient}, where the random physical
forces are used as forward propagations to reach new states. A more dedicated planner is the disassembly breadth first search (BFS)~\cite{tian2022assemble}, which uses physics-based force directions in combination with a breadth first search and state similarity checks. 

Our work extends approaches like MateVec-TRRT~\cite{Ebinger2018MateVecTRRT}, BK-RRT~\cite{zickler2009efficient}, and BFS~\cite{tian2022assemble} in that we focus on general removal tasks in disassembly scenarios, where we add the concept of scale-invariant sampling to ensure that we progress along long elongated narrow passages. 
Additionally, instead of using a possibly costly physical simulation~\cite{tian2022assemble}, we use purely geometrical arguments to find the direction of motion.

\subsection{Biased Sampling}

To circumnavigate narrow passages, most research has focused on biasing samples.
Common approaches include Obstacle-based
sampling~\cite{Boor1999GaussianSampling}, where samples are biased towards
obstacle boundaries, Bridge-based sampling~\cite{Hsu2003BridgeTest} where
samples are biased towards narrow passages, Utility-based
sampling~\cite{burns2005toward}, where samples are drawn based upon their
utility to improve the roadmap, and Dynamic-domain
RRT~\cite{yershova2005dynamic}, where the Voronoi bias of samples is adjusted
based on their performance. Sample biases can also be learned from similar
situations~\cite{ichter2018learning,chamzas2021learning} to improve sampling
quality.
Once paths are found, biasing can be used to improve the path quality by using
informed sampling~\cite{gammell2014informed}, where an admissible ellipsoid
heuristic is used in RRT*~\cite{karaman2011sampling} to bias samples towards paths with lower cost.
More recent approaches focus on selective densification~\cite{huang2025selective},
where sparse regions are identified, and their resolution is changed so that
more samples are produced in narrow passages. 
Our approach differs by concentrating on object extraction tasks while carefully finding trade-offs between uniform and scale-invariant sampling.

This trade-off is similar to the
exploration/exploitation trade-off~\cite{rickert2014balancing}, where
an exploration step finds locally free configuration space regions, while an
exploitation step uses this region to quickly find valid paths. 
We follow a similar approach, but see exploration as a search for high information entropy samples, while exploitation as the biasing operation along the high-information entropy samples.

\subsection{PCA-based sampling}

A widely used sampling analysis tool in motion planning is 
Principal component analysis (PCA)~\cite{pearson1901liii,jolliffe2011principal}.
PCA can extract a principal component from a set of samples, thereby showing a
direction in which the samples are distributed. This is particularly useful to find
directions to sample inside narrow passages. 

One way to leverage PCA is to detect if samples are currently inside a narrow passage.
For example, ADD-RRT~\cite{cai2022add} uses PCA to identify nodes which are likely in a narrow passages, and changes the sampling strategy accordingly in the immediate neighborhood.
If regions of interests are available, like in manipulation tasks, PCA can be
used to ensure sampling is better distributed in those
regions~\cite{rosell2013path}. Another application of PCA in this context is to
reduce the conformal states for protein folding~\cite{teodoro2003understanding}
thereby reducing the effective dimensions of the problem.

PCA can also be used globally to shape the sampling distributions around an existing
configuration. The pioneering work by Dalibard et
al.~\cite{dalibard2009control,dalibard2011linear} uses PCA to locally adjust the
sampling radius around a node proportional to the Eigenvalues along each
principal component. To make this more efficient, later work~\cite{lee2012sr}
used a bridge test~\cite{Hsu2003BridgeTest} to identify narrow passages before
running PCA on the samples. This reduces the calls to PCA to relevant regions of the configuration space.
For dynamical systems, PCA can also be used to learn the bias stemming from the
dynamics~\cite{li2010balancing} to make sampling more efficient.

Our work differs in that we do not consider PCA as a global tool for efficient
sampling~\cite{dalibard2011linear}, but instead leverage PCA as a local object extraction tool to exploit samples at high information entropy scales.

\subsection{Multi-Arm Bandit}

While novel sampling methods are important to better explore the space, it is
also important to decide when to use which sampling method.
A convenient framework to make this decision is the multi-arm bandit
(MAB)~\cite{auer2002finite,bubeck2012regret,slivkins2019introduction}. MABs have been used in
motion planning for different purposes, for example to select the best path from a set of candidate
paths~\cite{koval2015robust}, to select from different grasp strategies~\cite{eppner2017visual}, and to run multiple trees with different sampling
strategies, then select the best performing tree~\cite{Lai2022RRF}.

A major work in this area is the
MAB-RRT~\cite{faroni2023motion,faroni2024online}, which is an
RRT~\cite{Kuffner2000} with a MAB replacing the uniform sampling strategy. In the original MAB-RRT~\cite{faroni2023motion}, each arm is used to select a particular region of the state
space to sample. However, MAB-RRT is a general purpose planner which can take as input any set of sampling strategies if the MAB is clearly defined.
Our work is complementary in that we treat MABs, similar Faroni et
al.~\cite{faroni2023motion}, as our global decision framework. However, while we leverage MAB-RRT as our framework, we differ by tackling a different application (object extraction) while using different sampling strategies and reward functions in the MAB.

\section{Scale-Invariant Sampling in Multi-Arm Bandit Motion Planning}

\begin{algorithm}[t]
\def\Q{\mathcal{Q}}
\def\Vnew{\V_{\text{new}}}
\def\rmin{r_{\text{min}}}

\DontPrintSemicolon
\SetAlgoLined
\SetKwInOut{Input}{Input}
\SetKwInOut{Output}{Output}
\SetKwInOut{Parameters}{Parameters}
\caption{Finding High Information Entropy Scale}
\label{alg:adaptive-sphere}
\Input{start configuration $q_0$}
\Parameters{initial radius $r_0$; optimal validity rate interval
  $[\alpha_{\min},\alpha_{\max}]$; shrink factor $s$; growth factor
  $g$; min radius $\rmin$; batch size $b$, max steps $S$}
\Output{final radius $r^\star$, valid samples $\V$}

$r \gets r_0$;\quad $\V\gets \emptyset$ \;
\For{$i=1$ \KwTo $S$}{
  $\Q\gets$\textsc{SampleSphere}$(q_0, r, b)$ \;
  $\Vnew \gets \{q\in\Q \mid $\textsc{CheckMotion}$(q_0, q)\}$ \;
  $\alpha\gets \frac{|\Vnew|}{|\Q|}$ \tcp*{validity rate}

  \uIf{$\alpha \in [\alpha_{\min},\alpha_{\max}]$}{
    $r^\star\gets r$;\quad $\V\gets \V\cup \Vnew$;\quad \textbf{break}\;
  }
  \uElseIf{$\alpha < \alpha_{\min}$}{
    $r\gets r/s$;\quad $\V\gets \V\cup \Vnew$ \tcp*{shrink radius}
  }
  \Else{ $r\gets r\cdot g$ \tcp*{grow radius}}
}
\uIf{$r \le \rmin$}{ $r^\star\gets \rmin$ }
\Return $(r^\star,\V)$ \;
\end{algorithm}

\begin{algorithm}[t]
\DontPrintSemicolon
\SetAlgoLined
\SetKwInOut{Input}{Input}\SetKwInOut{Output}{Output}

\caption{Sample from PCA-Aligned Cylinder}
\label{alg:sample-pca-cylinder}

\Input{Normalized PCA axis $a_0$, direction $d \in \{0,1\}$, cylinder height interval $[h_{\min}, h_{\max}]$, extension $\delta$, radius $R$}
\Output{sample configuration $q$}

\tcp{(1) Sample random height vector}
$h \sim \mathcal{U}(h_{\min}, h_{\max})$\;
$a \gets h \cdot a_0$\;

\tcp{(2) Sample random direction in (N-1)-dimensional ball}
$u \sim \mathcal{U}(0, 1)$\;
$t \sim \mathcal{N}(\mathbf{0}, \mathbf{I}_{N-1})$\;
$p \gets R\cdot u^{1/(N-1)}$\tcp*{Account for volume density}
$b \gets p \cdot t / \|t\|$\;

\tcp{(3) Assemble vector to get final sample}
$Q \gets \textsc{OrthonormalBasis}(a)$\;
\Return $a + Q\cdot b$\;
\end{algorithm}

\begin{algorithm}[t]
\caption{Scale-Invariant MAB-RRT}
\label{alg:mabrrt}
\SetKwData{valid}{valid}
\DontPrintSemicolon
\SetAlgoLined

\KwIn{Start state $\xstart$, goal region $\Xgoal$}
\KwOut{Path $\pi$ or failure}

$(r^\star,\V) \gets$ \textsc{FindHighInfoEntropyScale}($x_{\text{start}}$)\;
$T \gets \{\xstart, \V\}$\;
{$\mathbf{a} \gets \textsc{ComputePCA}(\V)$} \tcp*{Principal escape direction}
$S \gets \{\textsc{Uniform}, \textsc{PC-Positive}({\mathbf{a},} r^\star), \textsc{PC-Negative}({\mathbf{a},} r^\star)\}$\;

\While{$\textbf{Not } \textsc{Terminate}()$}{
  {$h_{\text{ext}} \gets \delta \cdot r^\star$} \tcp*{Cylinder extension from $r^\star$}
  $s \gets \textsc{SelectArmMAB}(S)$ \tcp*{UCB arm selection}

  $x_{\text{sample}} \gets \textsc{Sample}(s, h_{\text{ext}})$\;
  $x_{\text{near}} \gets \textsc{Nearest}(T, x_{\text{sample}})$\;
  $x_{\text{new}} \gets \textsc{Steer}(x_{\text{near}}, x_{\text{sample}})$\;
  $\valid \gets \textsc{CollisionFree}(x_{\text{near}}, x_{\text{new}})$\;

  \If{$\valid$}{
    $T \gets T \cup \{x_{\text{near}}, x_{\text{new}}\}$\;
    \If{$s \neq \textsc{Uniform}$}{
      $\V \gets \V \cup \{x_{\text{new}}\}$ \tcp*{Accumulate valid samples}
      $\mathbf{a} \gets \textsc{RecomputePCA}(\V, \mathbf{a})$ \tcp*{Online recalibration}
      $r^\star \gets \max(r^\star,\, r_{\text{sample}})$ \tcp*{Expand reach}
    }
    \If{$x_{\text{new}} \in \Xgoal$}{
      \Return \textsc{ExtractPath}($T$, $x_{\text{new}}$)
    }
  }
  \textsc{UpdateReward}($S$, $s$, $\valid$)\;
}
\Return {$\failure$}\;
\end{algorithm}

We propose a new sampling scheme which consists of three interconnected methods:
Scale sampling to find high information entropy scales, directional sampling to
exploit those scales, and a multi-arm bandit planner to integrate those
samplers with classical sampling strategies. Each method is further detailed below.

\subsection{Exploration: Finding High Information Entropy Scales}

The scale sampler is an adaptive sampling strategy designed to identify radii
at which sampling yields a high information content. In particular, it aims to
find a sampling scale at which the fraction of valid samples lies within a
given target interval. This interval specifies a desired range of validity rates
that characterizes informative sampling scales.

The algorithm is depicted in Alg.~\ref{alg:adaptive-sphere}. 
At each iteration, the scale sampler draws a fixed-size batch of quasi-random
samples from a sphere (see below) of the current radius and evaluates their validity (Line 3,4).
Based on the observed validity rate, the radius is increased or
decreased (Line 6--12). This process continues until a radius is found whose validity rate
falls within the target interval (Line 6--7), or until a predefined computational budget of $S$ steps is exhausted. In that case, we check if the radius is smaller than the minimal radius, in which case we clip it (Line 14--15). Afterwards, the best radius so far is returned (Line 16). 

Importantly, the scale sampler does not attempt to compute an optimal
radius. Instead, it performs a feasibility-driven search whose
goal is to find any radius that yields informative samples under noisy evaluation with finite samples. Our method is thereby robust to stochastic variability
in validity estimates and avoids the need for strong assumptions such as
deterministic monotonicity of validity with respect to scale.

\paragraph{Sphere Sampling}

An important aspect of our algorithm is the sphere sampling scheme. When using uniform sampling, we often face clustering of points on the sphere, which negatively impacts the validity rate, which requires a more uniform coverage of samples. To achieve this, we employ a dual sampling scheme, where alternate between uniform sampling and a jittered Fibonacci lattice sampling scheme. The Fibonacci lattice provides low-discrepancy samples~\cite{Niederreiter1992,lavalle2006planning} which guarantee a better coverage that avoids clustering. Without jitter, the lattice would place samples at identical positions across radii, making coverage entirely dependent on the uniform component. This jitter is controlled by the parameter $p_{\text{fibo-jitter}}$.
%M_PI/8

%\paragraph{Difference to Bisection Search}

%A common approach for finding a minimum value along a range is Bisection search~\cite{kaw2009numerical}.
%However, this is not directly applicable to this case, since the problem is not a minimization problem. As mentioned above, we
%look for any radius whose validity rate lies within a given interval. Once such
%a radius is found, further refinement (as would be done by bisection search)
%is not relevant. 
%With bisection search, there is a clear risk of losing half of the
%search space due to an incorrect discard decision caused by noisy estimates, whereas we mitigate this risk using the grow/shrink strategy.
%
%Instead, we only observe noisy validity estimates at each selected radius
%based on finite samples. Each update step is therefore driven by stochastic
%feedback, while the update rule itself is deterministic, using multiplicative
%scaling to grow or shrink the radius. As a result, we do not require strict
%monotonicity at every step; as long as the grow and shrink parameters are
%not equal, the algorithm has the ability to explore both directions. 

\subsection{Exploitation: Principal Components Sampling}

Once a high-information entropy radius has been found, our planner should exploit it. A common way to better understand the valid sample distribution at the chosen scale is to run a principal component analysis (PCA). A PCA is a linear transformation of the data onto a new coordinate system, such that the (first) principal component (or principal axis) captures the largest variation in the data~\cite{bishop2006pattern}. PCA is a staple of many scientific software packages, and can be implemented with tools like Eigen~\cite{eigenweb}. 

However, to exploit the principal component for biased sampling requires a dedicated sampler. One possibility is to create a hyper-cylinder from the principal component and sample around it. This is depicted in  Alg.~\ref{alg:sample-pca-cylinder}. Our algorithm proceeds in three stages. First, we sample a random height from the axis $a_0$ to obtain a vector $a$ (Line 1--2). Second, we create an $(N-1)$-dimensional ball around $a$, which is orthogonal to $a$ itself. To achieve this, we first sample directly in an $(N-1)$-dimensional ball by getting a random distance variable $u$ (Line 3) and a random direction $t$ (Line 4). We then compute the radius $p$ by accounting for the volume density of the ball in $N-1$ dimensions (there is more density further out). We then take $p$ and use it to compute the final direction $b$ in the $N-1$ ball. Third, we assemble the final sample by projecting $b$ into the null-space of $a$. This is achieved by getting an orthonormal basis $Q$~\cite{axler2024linear} and projecting $b$ into it. The final result $a + Q\cdot b$ (Line 8) gives a uniform distributed sample inside the cylinder of radius $R$ around the principal component as desired.

\subsection{Integration of Samplers into Multi-Arm Bandit RRT}

To decide when to use which sampler and to integrate them with other sampling strategies, we utilize a geometric version of the multi-arm bandit RRT (MAB-RRT)~\cite{faroni2023motion}. The algorithm itself is depicted in Alg.~\ref{alg:mabrrt}. This planner is similar to RRT~\cite{Kuffner2000}, but uses in each iteration a possibly different sampling function as decided by the multi-arm bandit framework. This algorithm gets as input a start state $\xstart$, a goal region $\Xgoal$, and outputs a path $\pi$ or a failure. We start by searching for a useful (high-information entropy) scale (Line 1) using the grow-shrink algorithm (Alg.~\ref{alg:adaptive-sphere}). The outcome is a radius $r^{\star}$ and valid nodes and edges $V$. The nodes and edges are added to the initial tree at $\xstart$ (Line 2). PCA is then computed on $V$ to obtain the principal escape direction (Line~3). Afterwards, we define the set of samplers (Line~4), which includes a uniform sampler and the two principal component samplers, one for the positive and one for the negative direction. 

After those initialization steps, we start the inner loop while a terminate condition is not met. At each iteration, a cylinder extension height $h_{\text{ext}}$ is computed from the current best radius $r^\star$, allowing the principal component samplers to sample beyond the initial scale (see below). This inner loop starts with the selection of one sampler using the sliding window UCB policy denoted below. The remainder of the loop continues as in the original RRT by sampling a state, computing the nearest tree node, steering from this node to the sampled configuration, and checking if the connection is valid. If the connection is valid, we add it to the tree. For principal component samples, three additional updates occur: the valid sample is accumulated, the principal escape direction is recomputed online from all accumulated valid samples (see Online Recalibration below), and $r^\star$ is updated if the sample was drawn at a radius exceeding the current best, where $r_{\text{sample}}$ denotes the radius at which the cylinder sampler generated the point (see Cylinder Extension below). If both the connection is valid and the last added configuration belongs to the goal region, we extract the path from the tree through backward search and return it. If no solution has been found yet, we update the bandit rewards and restart the loop. If no solution has been found when the terminate condition becomes true, we end the loop and return a failure.   

\subsubsection{Multi-Arm Bandit Dynamics}

The multi-arm bandit algorithm selects samplers upon the Upper Confidence Bound (UCB) algorithm while each arm is updated via a reward function computed from the validity of a sample.

\paragraph{Arm Selection.} At each iteration, arm selection follows a sliding-window UCB policy~\cite{GarivierMoulines2008}:
\[
b^\star = \arg\max_{b \in B} \ \hat\mu_b + \beta \sqrt{\tfrac{\ln(\sum_c n_c + 1)}{n_b + 1}} ,
\]
where $b \in B$ are the bandit arms, $\hat\mu_b$ is the empirical mean rewards, $n_b$ is the recent count of arm $b$, $\beta$ is the exploration coefficient balancing exploitation against exploration, and $n_c$ is the count of the last $\parameterMABSlidingWindow$ arm iterations.
The window ensures that arm selection reflects recent performance, which is crucial in an object extraction task where the geometry changes as object parts separate. In practice, we scale uniform and sphere-based rewards by constants $\parameterUniformValidReward$ and $\parameterPCAValidReward$, which tune the trade-off between outward tree growth and narrow-passage exploration.

\paragraph{Reward function.}
Rewards are computed with respect to the last pulled arm, and the resulting value is propagated to all bandits along the decision path. Uniform samples are rewarded proportionally to their distance from the origin, encouraging outward tree expansion. Scale invariant samples (Positive and negative principal component) are rewarded inversely to their distance, prioritizing informative moves in narrow passages and down-weighting redundant samples once free space is reached. Invalid samples yield zero reward. Since the sliding-window UCB policy guarantees that each arm is selected infinitely often, and the uniform arm is itself probabilistically complete, the overall planner retains probabilistic completeness.

\paragraph{Online Recalibration.} During the planning loop, the principal component axis is not fixed to the initial estimate from the scale search. Each time a principal component sampler produces a valid sample, the principal component is recomputed. To prevent axis flipping due to the sign ambiguity of PCA, the new axis is checked for consistency with the previous axis via their dot product. This online recalibration allows the escape direction to adapt as the planner discovers more of the local geometry beyond the initial radius.

\paragraph{Cylinder Extension and Radius Growth.}

The principal component samplers update their sampling region through an integrated growth mechanism. Instead of setting the cylinder height to $r^\star$, an extension height of $h_{\text{ext}} = \delta \cdot r^\star$ with $\delta \geq 0$ being the extension factor. 
With $\delta = 0$, no extension occurs and samples are drawn at exactly $r^\star$ along the axis. Each principal component sampler samples along its respective direction within the range $[r^\star, r^\star + h_{\text{ext}}]$ from the origin along the axis.
Importantly, $r^\star$ is not fixed after the scale search: whenever a valid principal component sample is generated at a radius exceeding the current $r^\star$, the radius is updated. Since $h_{\text{ext}}$ is recomputed from $r^\star$, the escape direction is further increased for subsequent iterations.

\subsection{Open Source Software Implementation}

The samplers and the multi-arm bandit RRT have been implemented as an extension of the open motion planning library (OMPL)~\cite{sucan2012the-open-motion-planning-library}. This code is open source and available on Github\footnote{Link: \url{https://github.com/serboba/ompl/tree/MAB-RRT}}. We added a MAB-RRT planner demo, which can run on arbitrary occupancy maps in 2D. 
\begin{figure}[t]
    \centering
    \begin{subfigure}[b]{0.325\linewidth}
        \centering
        \includegraphics[width=\linewidth]{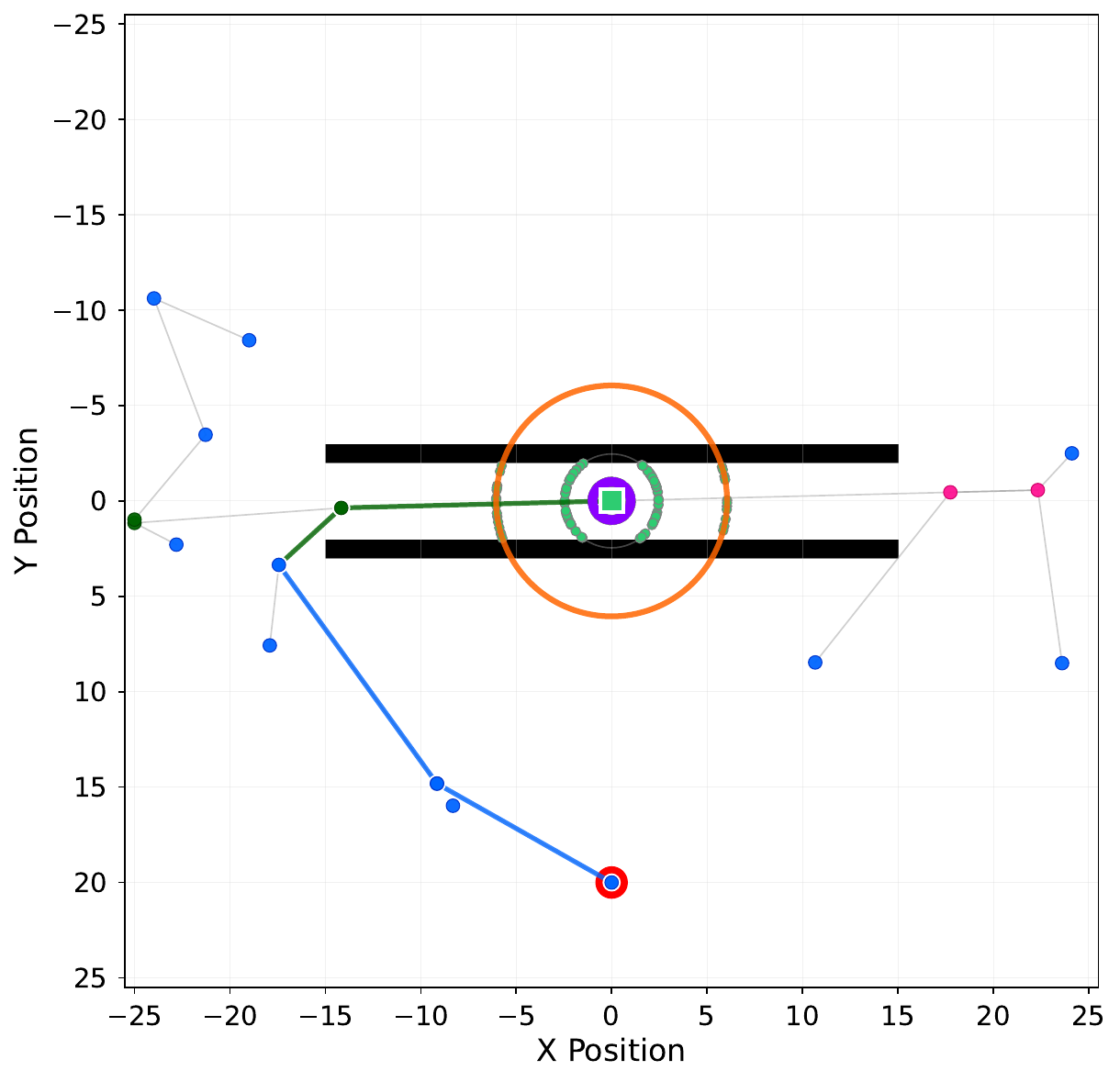}
        \caption{Tunnel experiment 1}
    \end{subfigure}
    \hfill
    \begin{subfigure}[b]{0.325\linewidth}
        \centering
        \includegraphics[width=\linewidth]{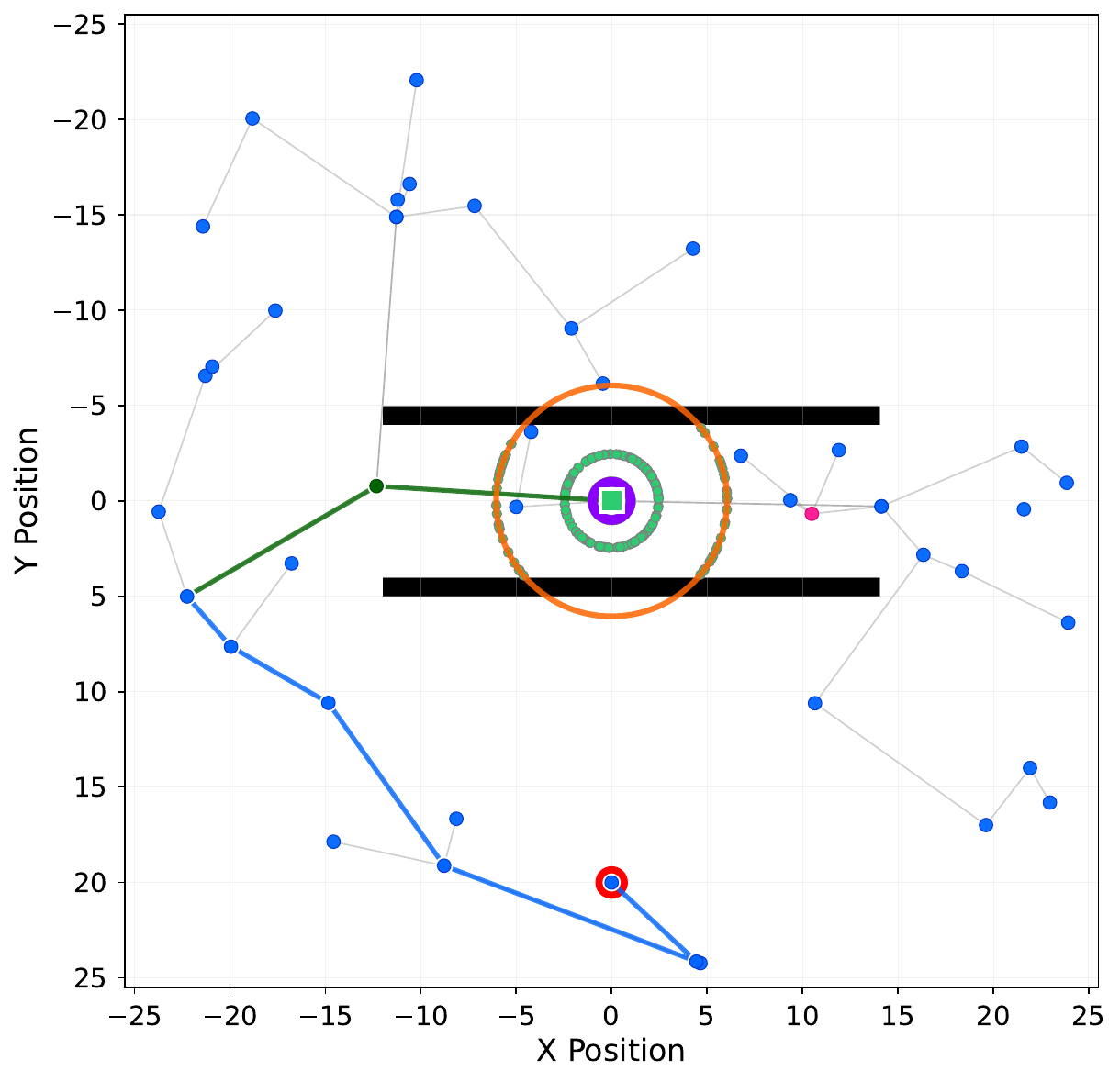}
        \caption{Tunnel experiment 2}
    \end{subfigure}
    \hfill
    \begin{subfigure}[b]{0.325\linewidth}
        \centering
        \includegraphics[width=\linewidth]{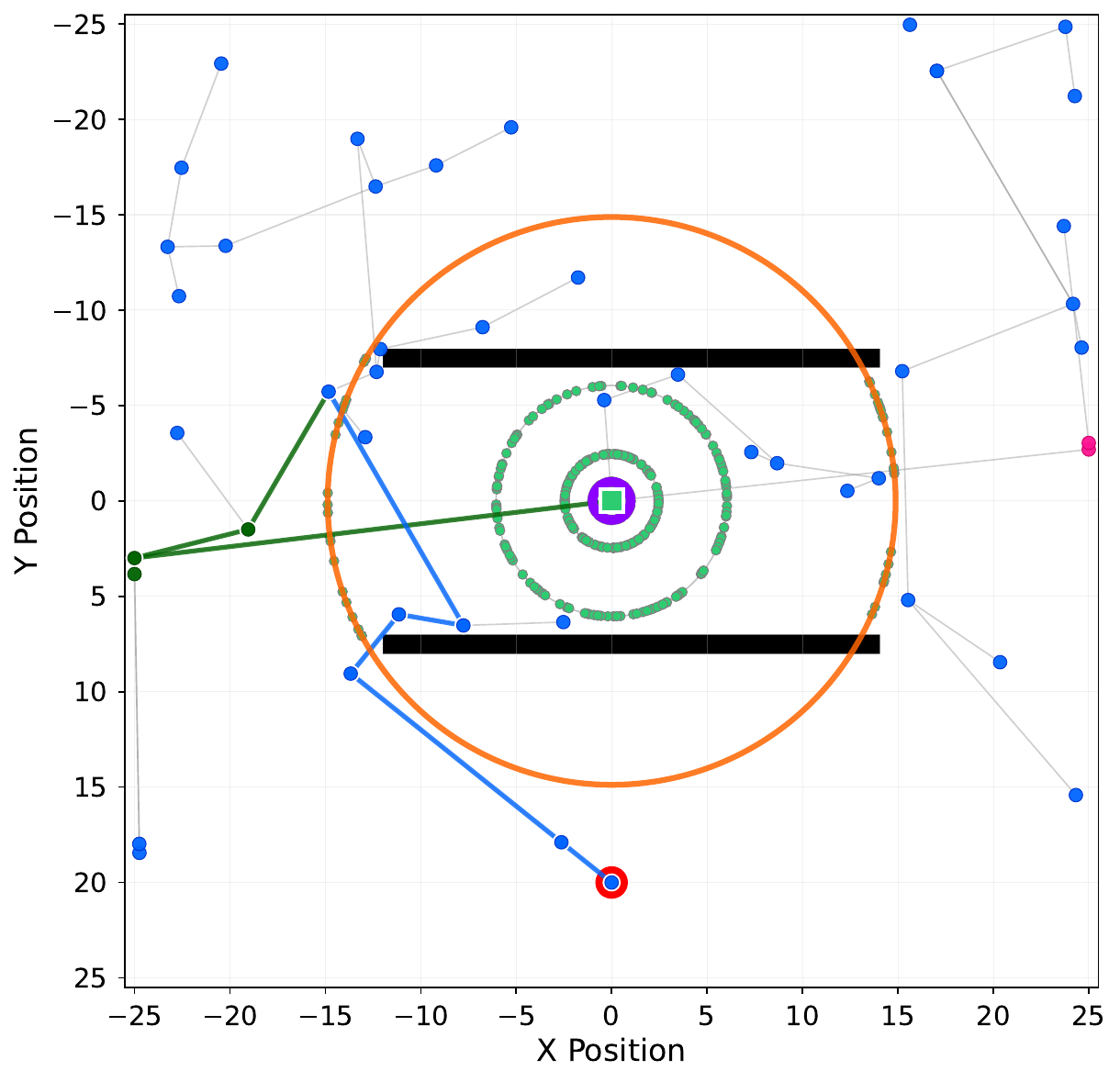}
        \caption{Tunnel experiment 3}
    \end{subfigure}

    \caption{Visualization of scale-invariant sampling in tunnel passages of different sizes (5 units, 10 units, and 15 units).
    Final burn-in radius is shown as \legendline{orange}, and path are shown depending on the type of active sampler:
    \legendline{blue} uniform, \legendline{green!60!black} positive principal component,
    and \legendline{magenta} negative principal component.
    Start is marked with \csquare{green!70},
    goal with \cmark{red}.
    Valid samples are shown as \cbox{blue} uniform,
    \cbox{green!60!black} positive principal component,
    \cbox{magenta} negative principal component,
    and burn-in valid samples as \cbox{green!70}.}
    \label{fig:toy-environments}
\end{figure}

\subsection{Example: Tunnel Environment}

To showcase scale-invariant sampling with MAB-RRT, we created three toy scenarios as shown in Fig.~\ref{fig:toy-environments}. Those three scenarios are tunnel environments, where a point robot has to move from the origin (green point) to a goal configuration (red point). The scenarios differ by the size of a tunnel obstacle (black bars), where two bars restrict the start configuration from above and below. The distance between the bars differ for each scenario (5 units, 10 units, and 15 units).

For each scenario, we visualize different properties of scale-invariant sampling. First, we show the final burn-in radius (high information entropy), at which roughly half of the samples are valid and half invalid. Furthermore, we show the search trees and the way the samples have been obtained. This includes uniform sampling (blue), positive principal component (green), and negative principal component (magenta). It can be seen that the high information entropy radius gives a good insight into where to move in each scenario, which leads to an efficient exploration of the configuration space. 
\begin{figure}[!ht]
    \centering
    % Global parameters: tune these fractions if needed
    % All images will have exactly the same width and height
    \newcommand{\imgwidthfraction}{0.23}
    \newcommand{\imgheightfraction}{0.23}

    % Derived dimensions (as fractions of \linewidth)
    \newlength{\imgwidth}
    \setlength{\imgwidth}{\imgwidthfraction\linewidth}
    \newlength{\imgheight}
    \setlength{\imgheight}{\imgheightfraction\linewidth}

    % Subfigure width: a bit larger than image to leave room for caption
    \newlength{\subfigwidth}
    \setlength{\subfigwidth}{1.05\imgwidth}

    % Row 1
    \begin{subfigure}[b]{\subfigwidth}
        \centering
        \includegraphics[width=\imgwidth, height=\imgheight, keepaspectratio]{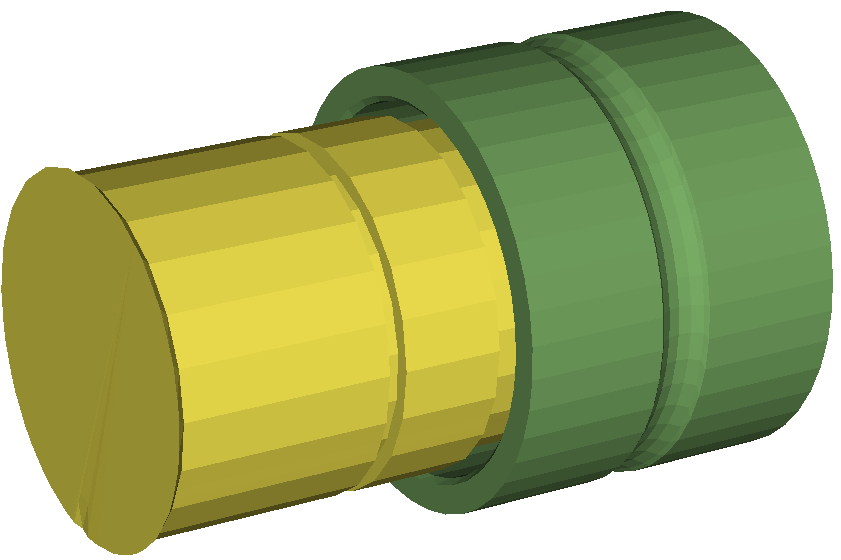}
        \caption{Socket: Start}
    \end{subfigure}
    \hfill
    \begin{subfigure}[b]{\subfigwidth}
        \centering
        \includegraphics[width=\imgwidth, height=\imgheight, keepaspectratio]{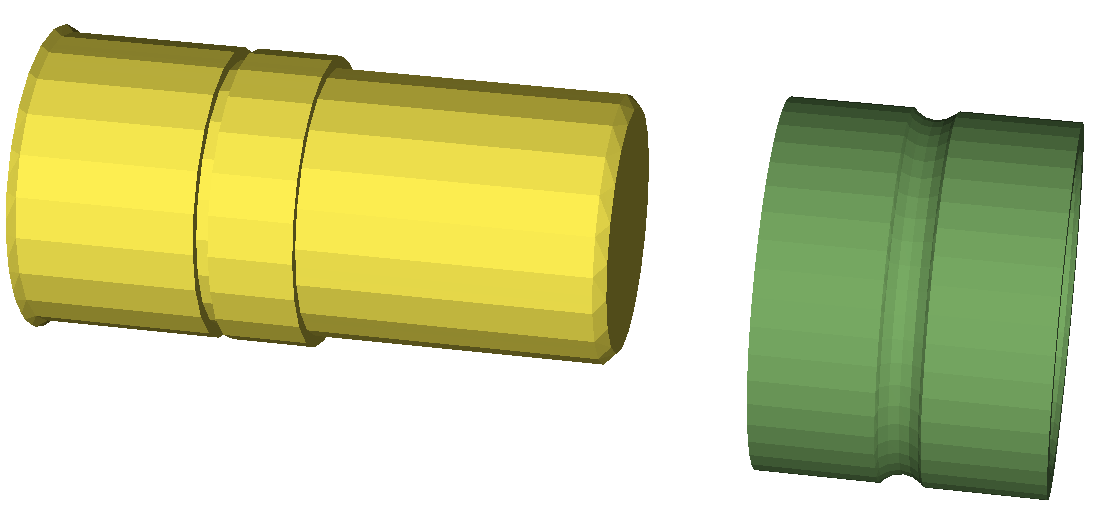}
        \caption{Socket: End}
    \end{subfigure}
    \hfill
    \begin{subfigure}[b]{\subfigwidth}
        \centering
        \includegraphics[width=\imgwidth, height=\imgheight, keepaspectratio]{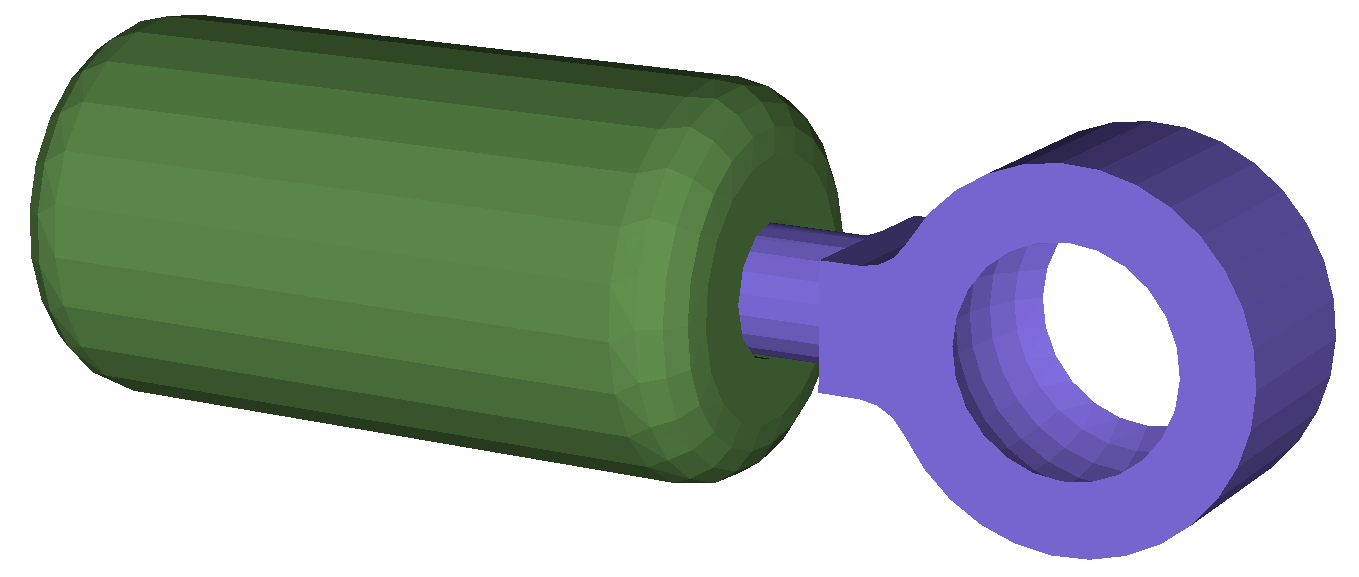}
        \caption{Eye Bolt: Start}
    \end{subfigure}
    \hfill
    \begin{subfigure}[b]{\subfigwidth}
        \centering
        \includegraphics[width=\imgwidth, height=\imgheight, keepaspectratio]{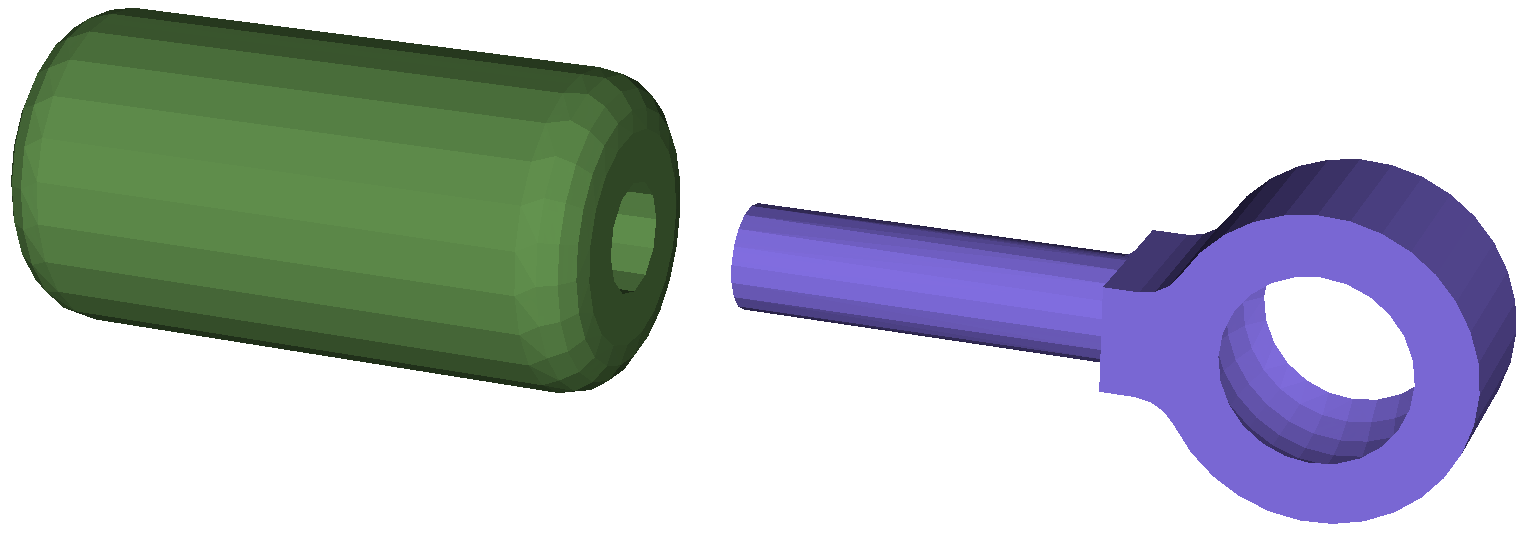}
        \caption{Eye Bolt: End}
    \end{subfigure}

    \vspace{1em}

    % Row 2
    \begin{subfigure}[b]{\subfigwidth}
        \centering
        \includegraphics[width=\imgwidth, height=\imgheight, keepaspectratio]{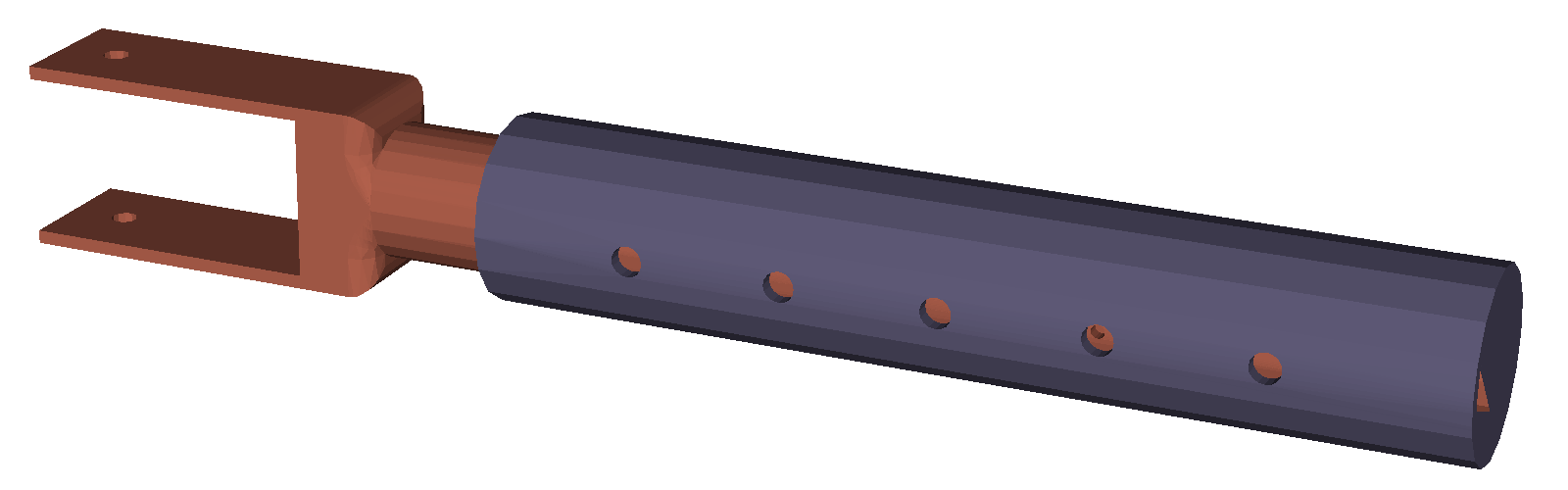}
        \caption{U-Bolt: Start}
    \end{subfigure}
    \hfill
    \begin{subfigure}[b]{\subfigwidth}
        \centering
        \includegraphics[width=\imgwidth, height=\imgheight, keepaspectratio]{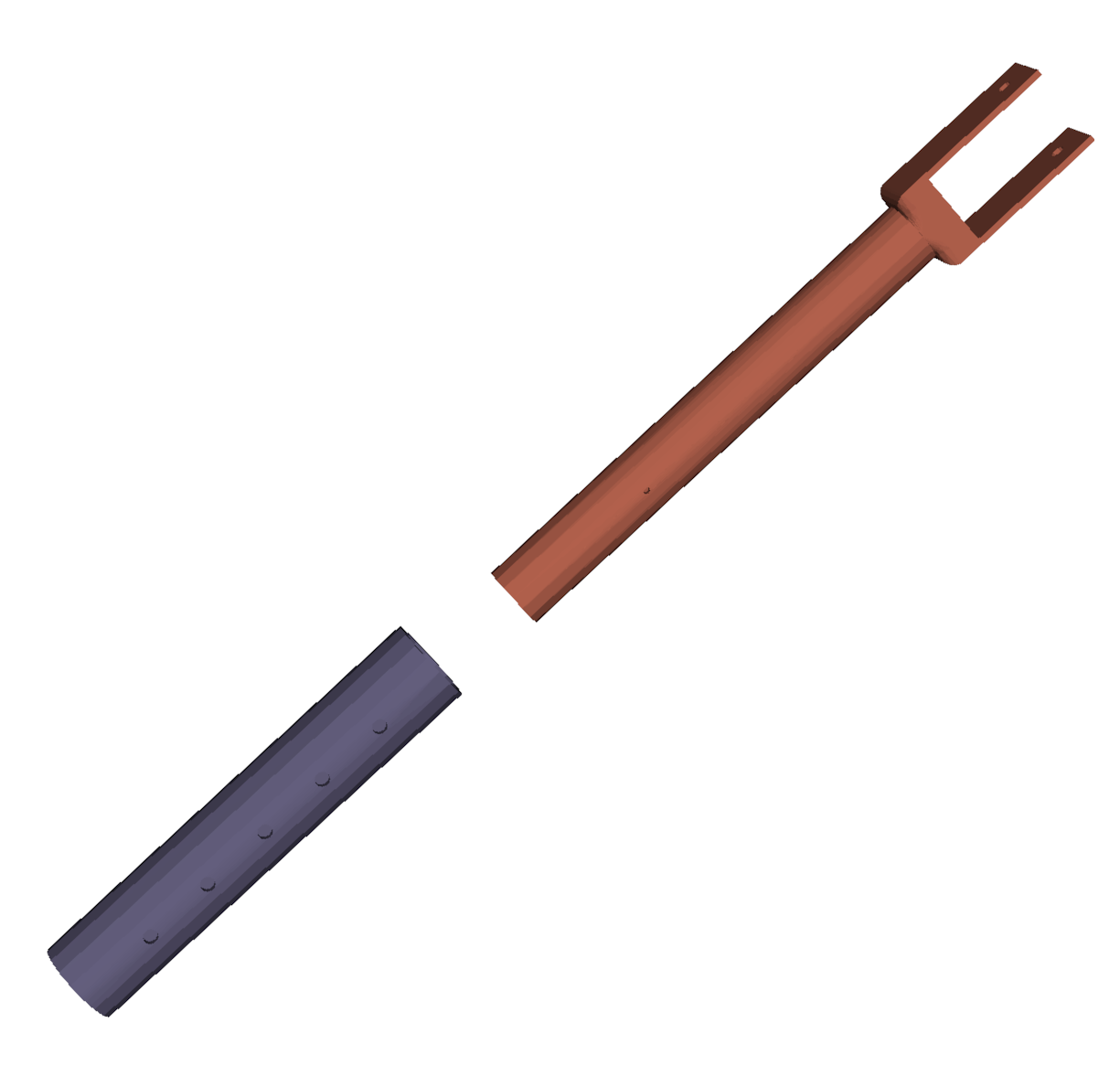}
        \caption{U-Bolt: End}
    \end{subfigure}
    \hfill
    \begin{subfigure}[b]{\subfigwidth}
        \centering
        \includegraphics[width=\imgwidth, height=\imgheight, keepaspectratio]{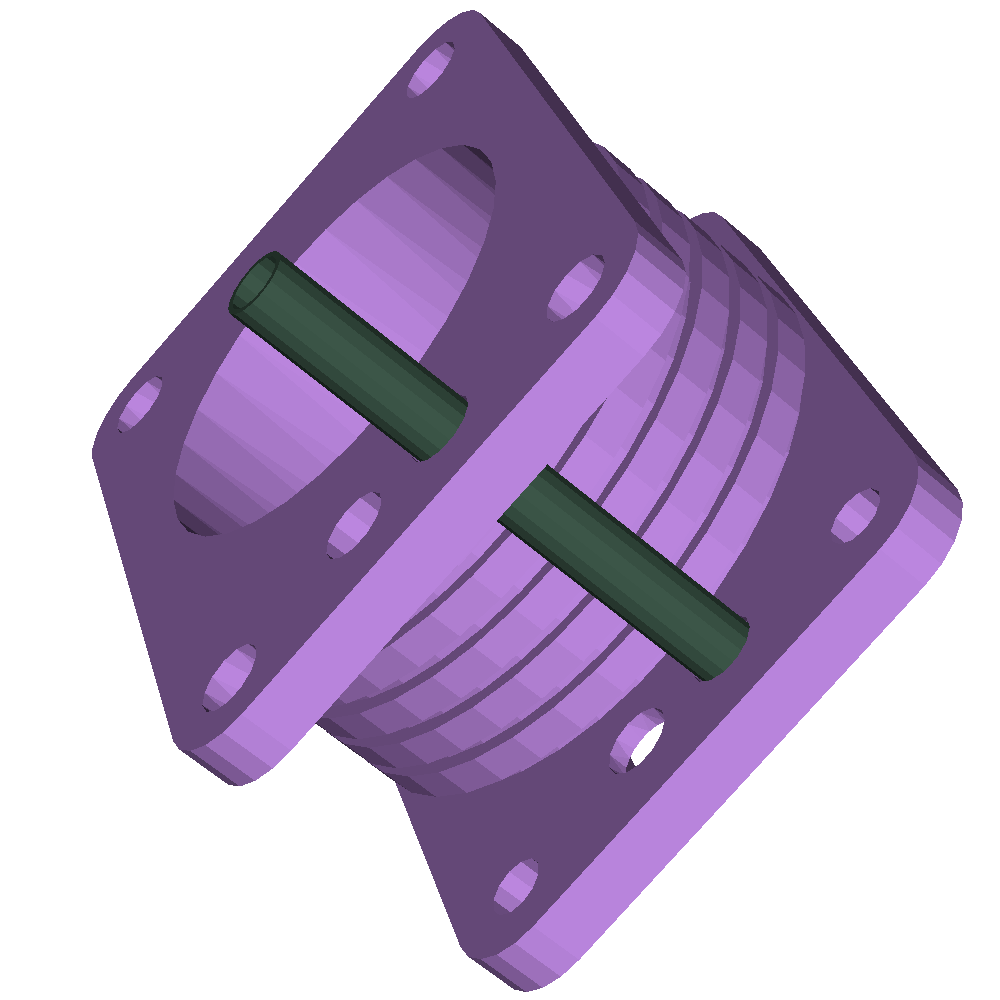}
        \caption{Motor Flange: Start}
    \end{subfigure}
    \hfill
    \begin{subfigure}[b]{\subfigwidth}
        \centering
        \includegraphics[width=\imgwidth, height=\imgheight, keepaspectratio]{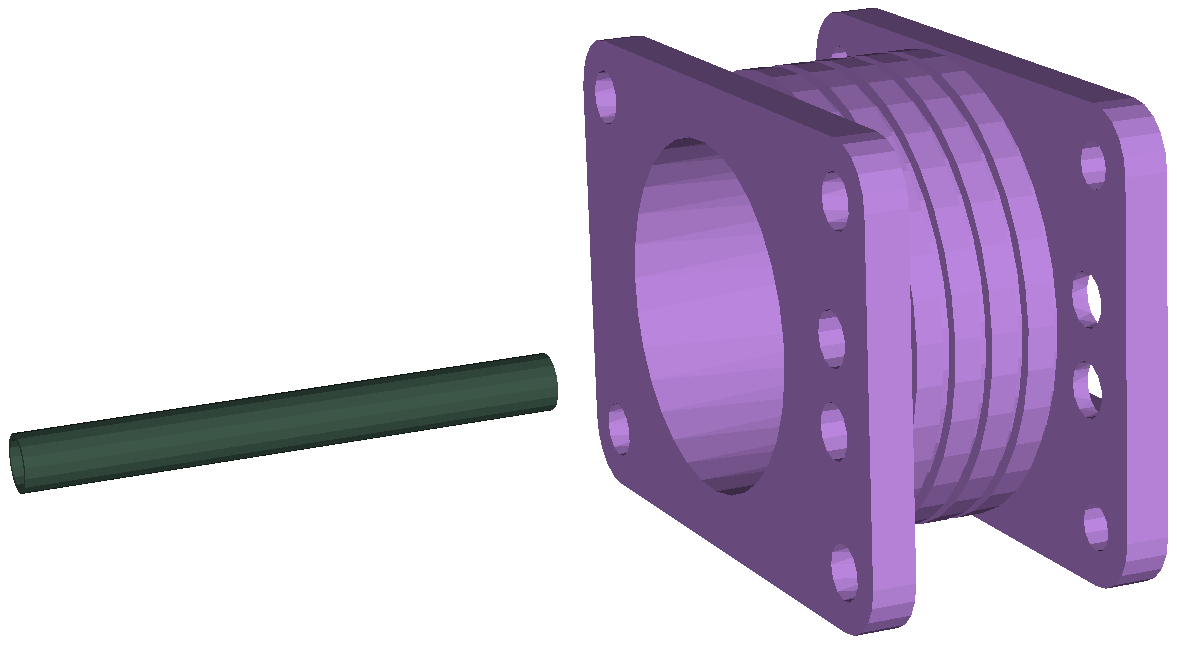}
        \caption{Motor Flange: End}
    \end{subfigure}

    \vspace{1em}

    % Row 3
    \begin{subfigure}[b]{\subfigwidth}
        \centering
        \includegraphics[width=\imgwidth, height=\imgheight, keepaspectratio]{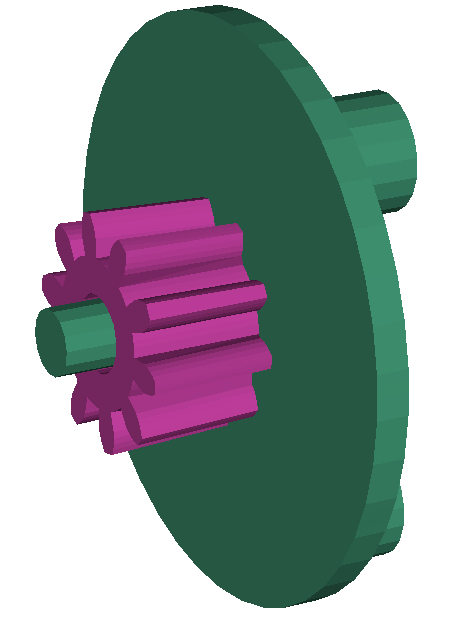}
        \caption{Gear Reducer: Start}
    \end{subfigure}
    \hfill
    \begin{subfigure}[b]{\subfigwidth}
        \centering
        \includegraphics[width=\imgwidth, height=\imgheight, keepaspectratio]{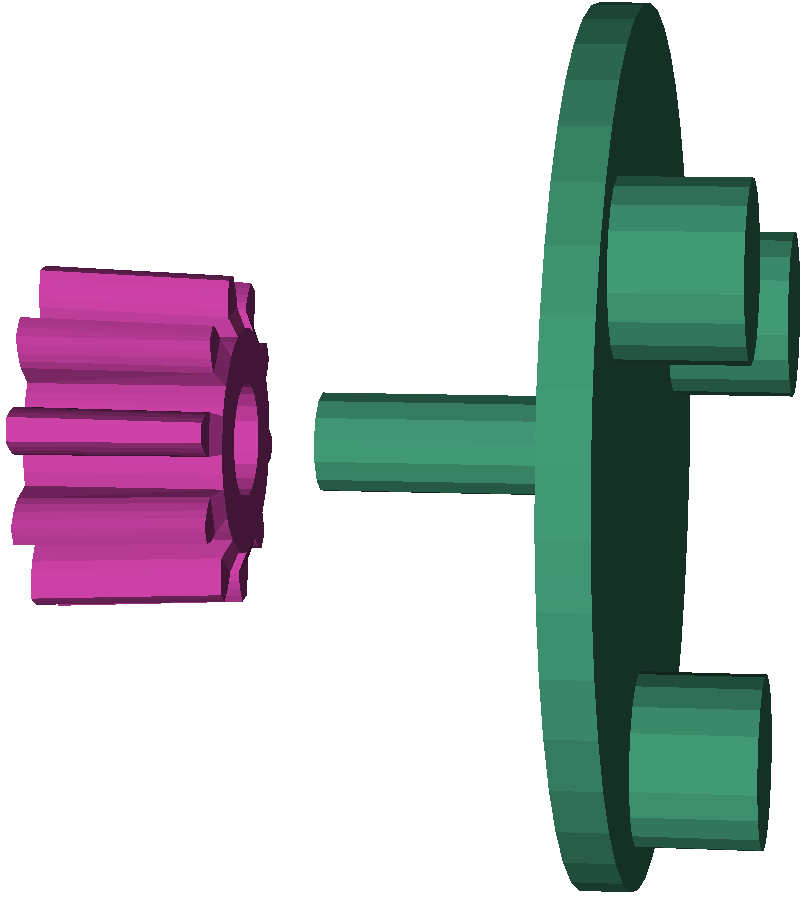}
        \caption{Gear Reducer: End}
    \end{subfigure}
    \hfill
    \begin{subfigure}[b]{\subfigwidth}
        \centering
        \includegraphics[width=\imgwidth, height=\imgheight, keepaspectratio]{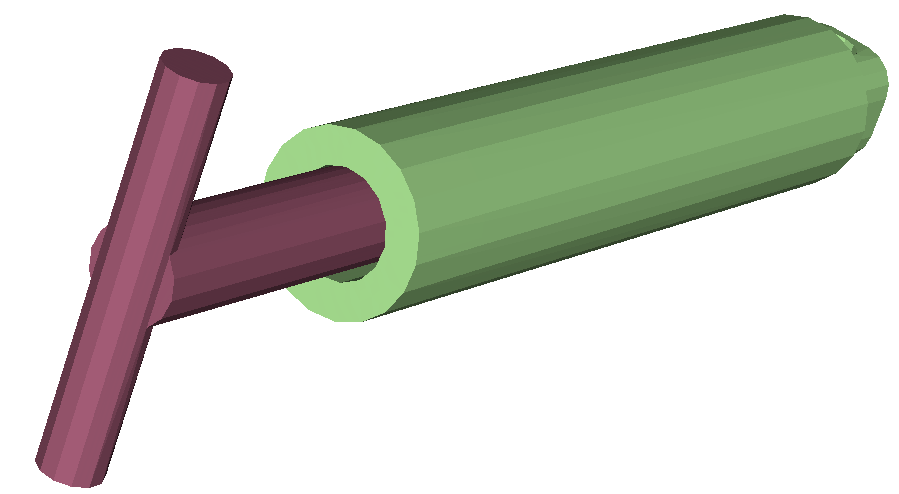}
        \caption{T-Bolt: Start}
    \end{subfigure}
    \hfill
    \begin{subfigure}[b]{\subfigwidth}
        \centering
        \includegraphics[width=\imgwidth, height=\imgheight, keepaspectratio]{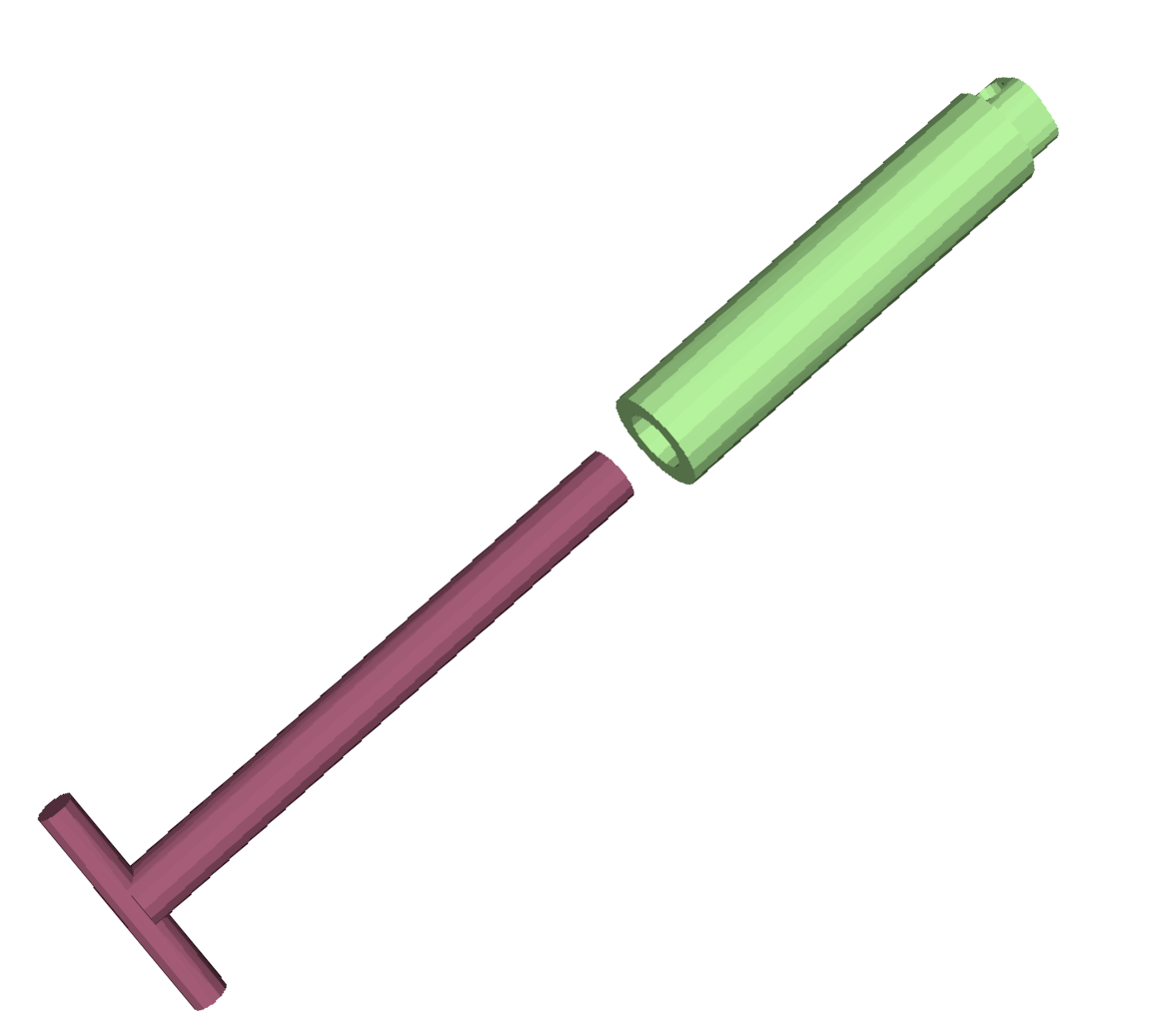}
        \caption{T-Bolt: End}
    \end{subfigure}

    \vspace{1em}

    % Row 4
    \begin{subfigure}[b]{\subfigwidth}
        \centering
        \includegraphics[width=\imgwidth, height=\imgheight, keepaspectratio]{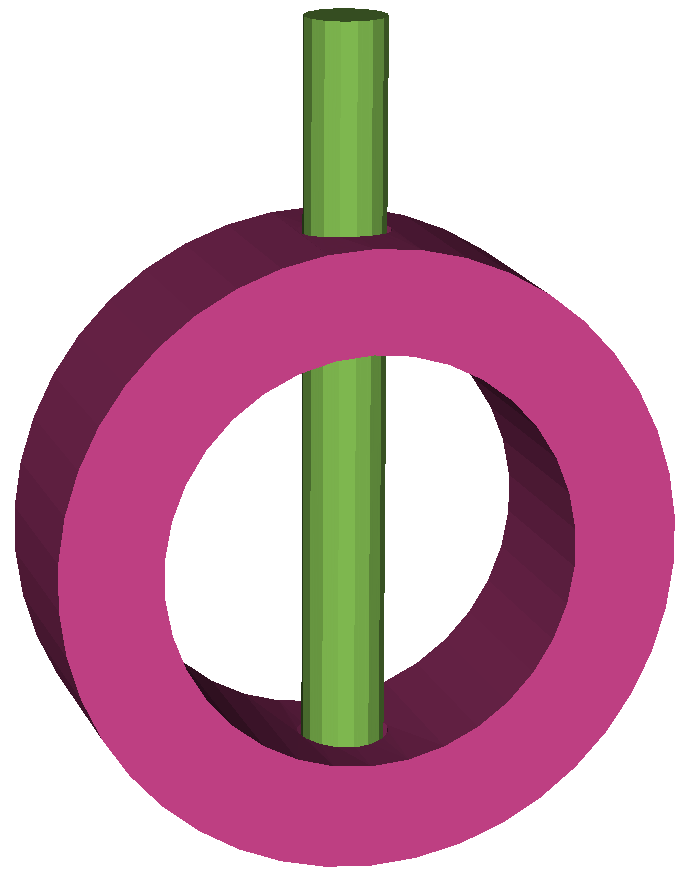}
        \caption{Ring Bolt: Start}
    \end{subfigure}
    \hfill
    \begin{subfigure}[b]{\subfigwidth}
        \centering
        \includegraphics[width=\imgwidth, height=\imgheight, keepaspectratio]{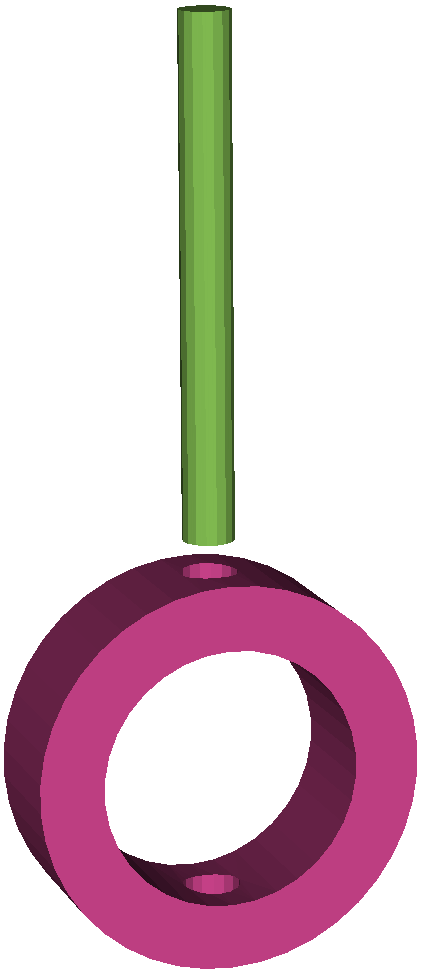}
        \caption{Ring Bolt: End}
    \end{subfigure}
    \hfill
    \begin{subfigure}[b]{\subfigwidth}
        \centering
        \includegraphics[width=\imgwidth, height=\imgheight, keepaspectratio]{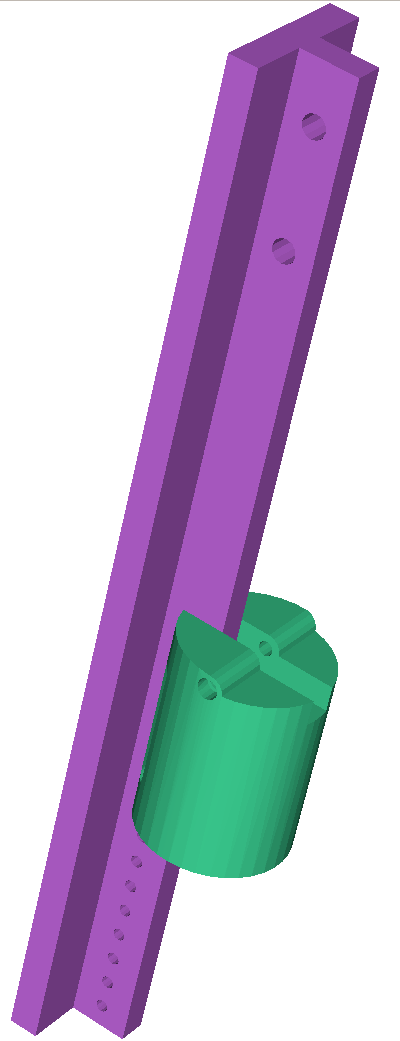}
        \caption{Cross-Pin Connector: Start}
    \end{subfigure}
    \hfill
    \begin{subfigure}[b]{\subfigwidth}
        \centering
        \includegraphics[width=\imgwidth, height=\imgheight, keepaspectratio]{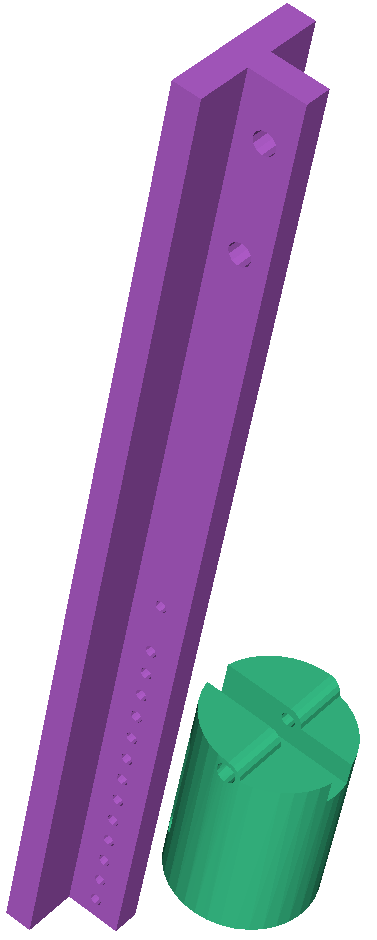}
        \caption{Cross-Pin Connector: End}
    \end{subfigure}
    \caption{Environment Comparison: Start vs. End states for selected environments. Each pair shows the initial (left) and final (right) configuration.}
    \label{fig:env-comparison}
\end{figure}
\begin{figure}[!ht]
    \centering

    % Global parameters (adjust these once to affect all subfigures)
    \def\plotheight{0.5\linewidth}   % Desired maximum height for all plots
    \def\plottoptrim{0.6cm}         % Amount to crop from the top (e.g., change to 2cm)
    \def\plotvspace{0.2em}
    \def\subWidth{0.95\textwidth}
    % --- Row 1 ---
    \begin{subfigure}[b]{0.49\textwidth}
        \centering
        \includegraphics[width=\subWidth,trim=0 0 0 \plottoptrim,clip]{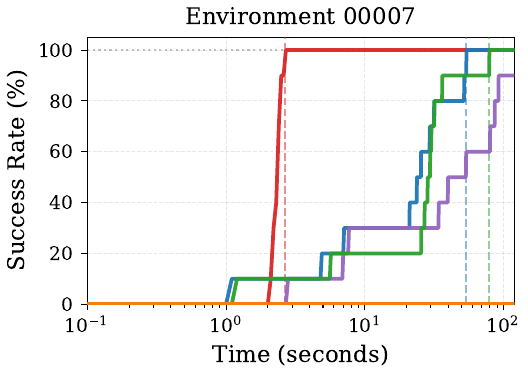}
        \caption{Socket}
    \end{subfigure}
    \hfill
    \begin{subfigure}[b]{0.49\textwidth}
        \centering
        \includegraphics[width=\subWidth,trim=0 0 0 \plottoptrim,clip]{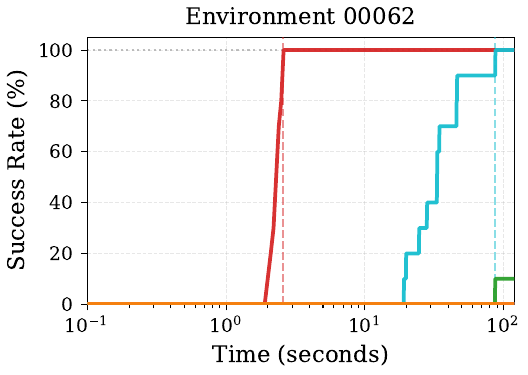}
        \caption{Eye Bolt}
    \end{subfigure}

    \vspace{\plotvspace} % Space between rows

    % --- Row 2 ---
    \begin{subfigure}[b]{0.49\textwidth}
        \centering
        \includegraphics[width=\subWidth,trim=0 0 0 \plottoptrim,clip]{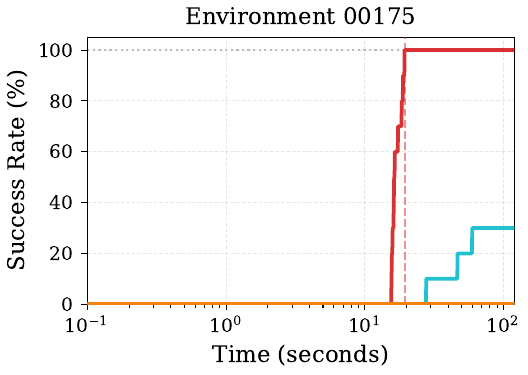}
        \caption{U-Bolt}
    \end{subfigure}
    \hfill
    \begin{subfigure}[b]{0.49\textwidth}
        \centering
        \includegraphics[width=\subWidth,trim=0 0 0 \plottoptrim,clip]{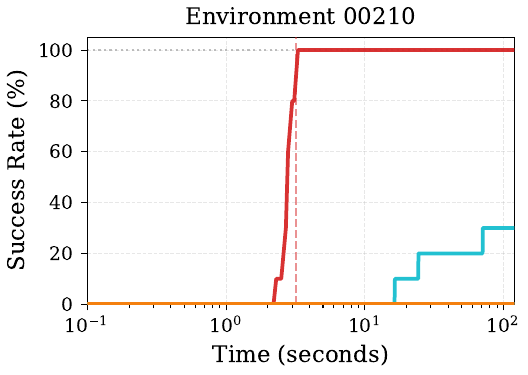}
        \caption{Motor Flange}
    \end{subfigure}

    \vspace{\plotvspace} % Space between rows

    % --- Row 3 ---
    \begin{subfigure}[b]{0.49\textwidth}
        \centering
        \includegraphics[width=\subWidth,trim=0 0 0 \plottoptrim,clip]{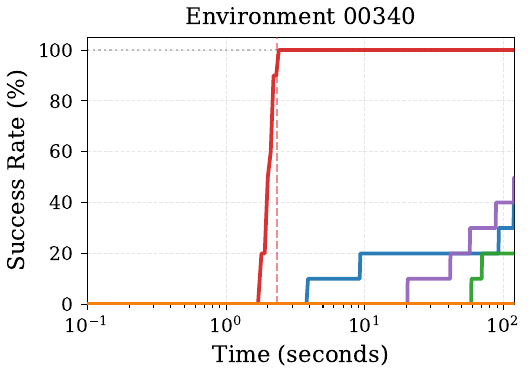}
        \caption{Gear Reducer}
    \end{subfigure}
    \hfill
    \begin{subfigure}[b]{0.49\textwidth}
        \centering
        \includegraphics[width=\subWidth,trim=0 0 0 \plottoptrim,clip]{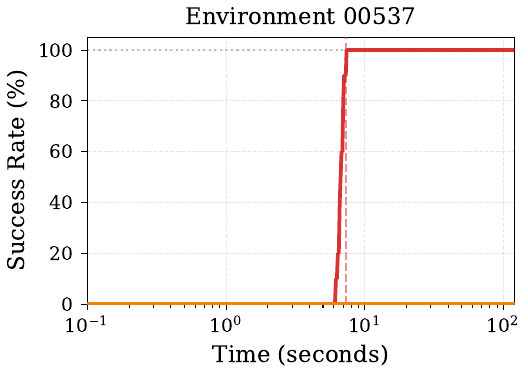}
        \caption{T-Bolt}
    \end{subfigure}

    \vspace{\plotvspace} % Space between rows

    % --- Row 4 ---
    \begin{subfigure}[b]{0.49\textwidth}
        \centering
        \includegraphics[width=\subWidth,trim=0 0 0 \plottoptrim,clip]{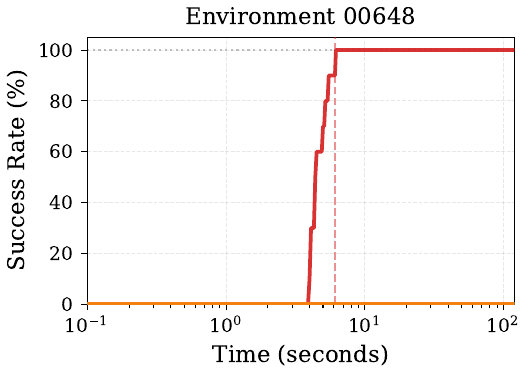}
        \caption{Ring Bolt}
    \end{subfigure}
    \hfill
    \begin{subfigure}[b]{0.49\textwidth}
        \centering
        \includegraphics[width=\subWidth,trim=0 0 0 \plottoptrim,clip]{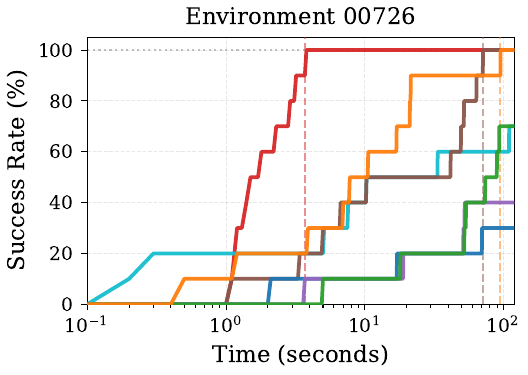}
        \caption{Cross-Pin Connector}
    \end{subfigure}

  \caption{Comparison of success rate using planners 
  \cbox{mabrrtcolor} MAB-RRT (ours),   
  \cbox{bridgerrtcolor} RRT + Bridge Sampling~\cite{Hsu2003BridgeTest},          
  \cbox{gaussianrrtcolor} RRT + Gaussian Sampling~\cite{Boor1999GaussianSampling}, 
  \cbox{obstaclerrtcolor} RRT + Obstacle Sampling~\cite{Amato1998ObstacleBased},
  \cbox{matevectrrtcolor} MateVec-TRRT~\cite{Ebinger2018MateVecTRRT},
  \cbox{bkrrtcolor} BK-RRT~\cite{zickler2009efficient,tian2022assemble}, 
  \cbox{bfscolor} BFS~\cite{tian2022assemble}, 
  \label{fig:results}} 
    
\end{figure}

\section{Results}

We evaluate our scale-invariant sampler on $8$ environments involving multiple
tight narrow passage scenarios inspired by realistic disassembly problems. The eight environments are shown in Fig.~\ref{fig:env-comparison} and represent a diverse set of disassembly tasks adopted from an existing dataset~\cite{tian2022assemble}. 

\paragraph{Experimental Setup}

All experiments were conducted on a laptop running Ubuntu 20.04.6 LTS with an Intel Core i7-5820K CPU (3.30GHz, 12 cores) and 32GB RAM. In each scenario, we compare MAB-RRT using scale-invariant sampling with six
alternative sampling strategies. This includes two sets of planners. First a set
of classical sampling strategies integrated into RRT~\cite{Kuffner2000}, namely
RRT + bridge sampling~\cite{Hsu2003BridgeTest}, RRT + Gaussian
sampling~\cite{Boor1999GaussianSampling}, and RRT + obstacle-based
sampling~\cite{Amato1998ObstacleBased}. Second, a set of modern strategies,
including mating vectors (MateVec-TRRT)~\cite{Ebinger2018MateVecTRRT}, physical simulation using behavioral kinodynamic RRT (BK-RRT)~\cite{zickler2009efficient,tian2022assemble}, and physical simulation using similarity checks and disassembly breadth first search (BFS)~\cite{tian2022assemble}. For each scenario, we run each planner for a total of $10$ runs with a timeout of $100$ seconds.

\paragraph{Hardware and Parameters}
For MAB-RRT, we use the following parameter values. For scale-invariant sampling, we use a Fibonacci jitter $\parameterFiboJitter = \parameterFiboJitterValue$, initial radius $\parameterScaleSamplingInitialRadius = \parameterScaleSamplingInitialRadiusValue$, min radius $\parameterScaleSamplingMinRadius = \parameterScaleSamplingMinRadiusValue$, max radius $\parameterScaleSamplingMaxRadius = \parameterScaleSamplingMaxRadiusValue$, batch size 
$\parameterScaleSamplingBatchSize=\parameterScaleSamplingBatchSizeValue$,
growth factor
$\parameterScaleSamplingGrowthFactor = \parameterScaleSamplingGrowthFactorValue$,
shrink factor
$\parameterScaleSamplingShrinkFactor = \parameterScaleSamplingShrinkFactorValue$,
max steps
$\parameterScaleSamplingMaxSteps = \parameterScaleSamplingMaxStepsValue$,
and validity rates 
$\parameterScaleSamplingMinValidityRate=\parameterScaleSamplingMinValidityRateValue$ and $\parameterScaleSamplingMaxValidityRate=\parameterScaleSamplingMaxValidityRateValue$.
For the Multi-arm bandit, we use a sliding window of $\parameterMABSlidingWindow = \parameterMABSlidingWindowValue$, an UCB exploration coefficient $\beta = \sqrt{2}$, and constants for the uniform rewards of $\parameterUniformValidReward = \parameterUniformValidRewardValue$ and PCA samplers as $\parameterPCAValidReward=\parameterPCAValidRewardValue$.
For the classical sampling strategies, we use the default parameters as specified in OMPL~\cite{sucan2012the-open-motion-planning-library,moll2015benchmarking-motion-planning-algorithms}. For the modern strategies, we use the default parameters as specified by the Assemble-Them-All~\cite{tian2022assemble} software package~\footnote{Link: \url{https://github.com/yunshengtian/Assemble-Them-All}}. 

\begin{figure}[htbp]
    \centering
    \setlength{\lineskip}{0.5em}
    \begin{minipage}[t]{0.33\linewidth}
        \centering
        \includegraphics[width=0.90\linewidth]{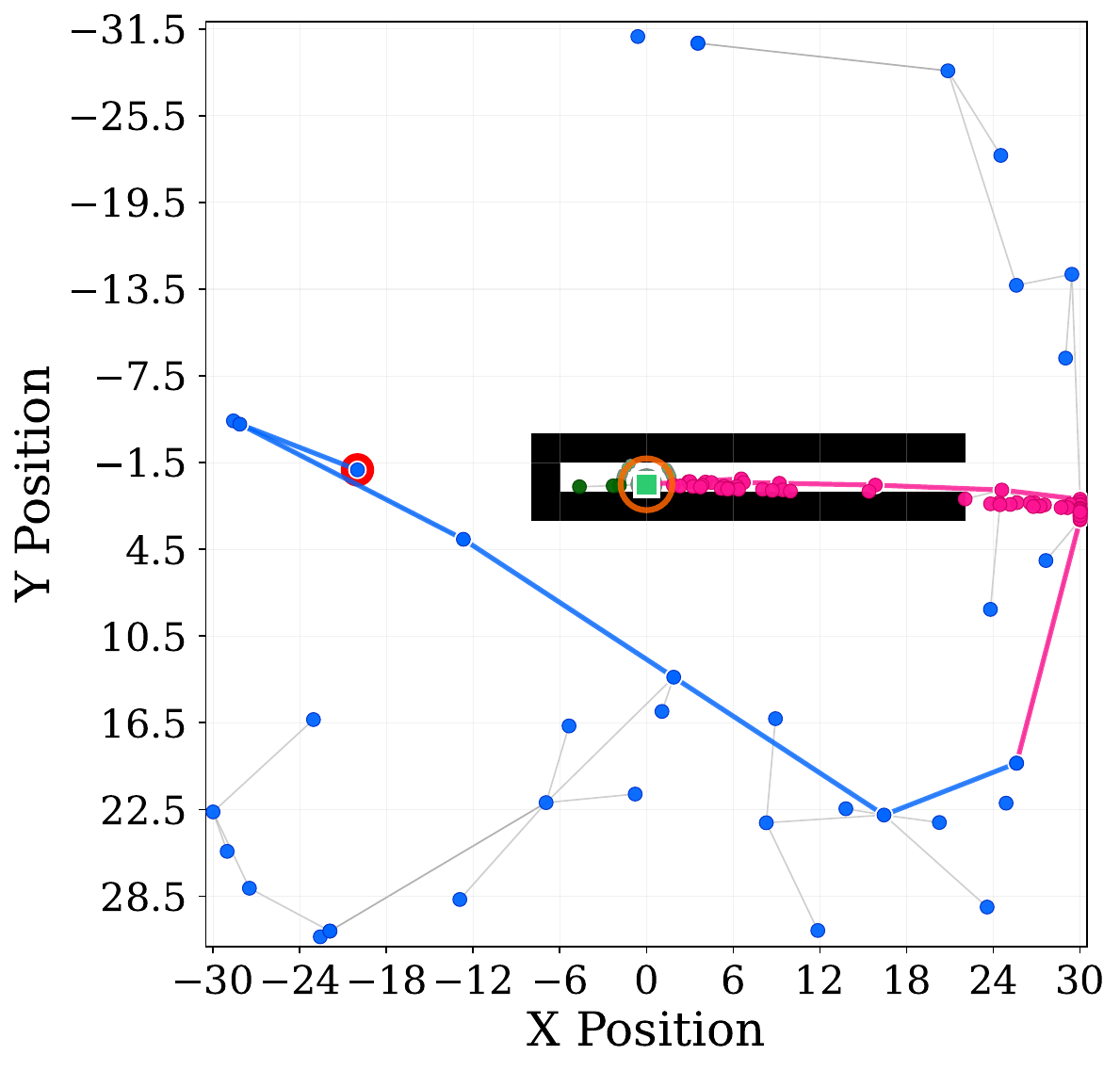} \par
        \includegraphics[width=0.90\linewidth]{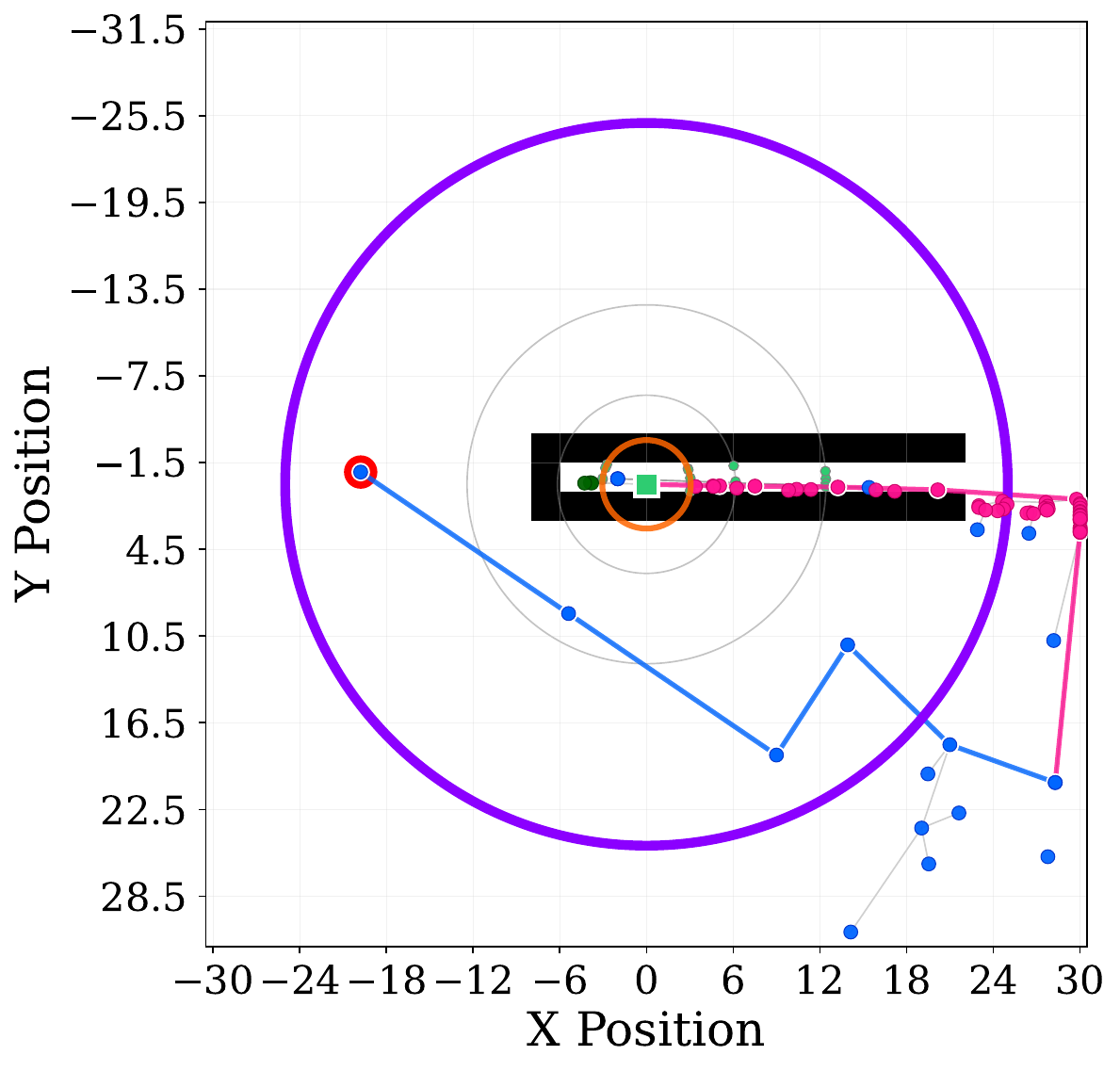} \par
    \end{minipage}%
    \hfill
    \begin{minipage}[t]{0.33\linewidth}
        \centering
        \includegraphics[width=0.90\linewidth]{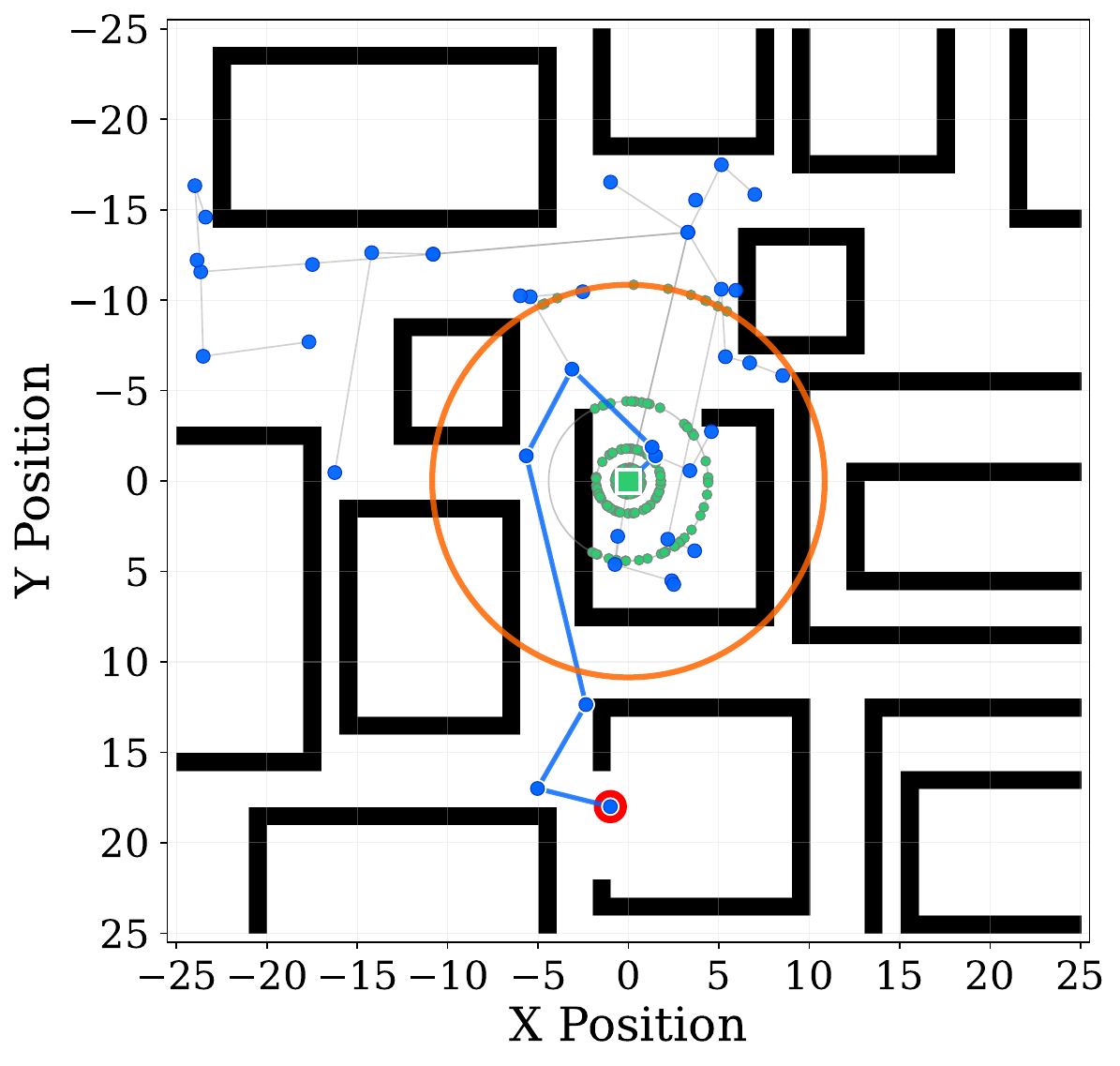} \par
        \includegraphics[width=0.90\linewidth]{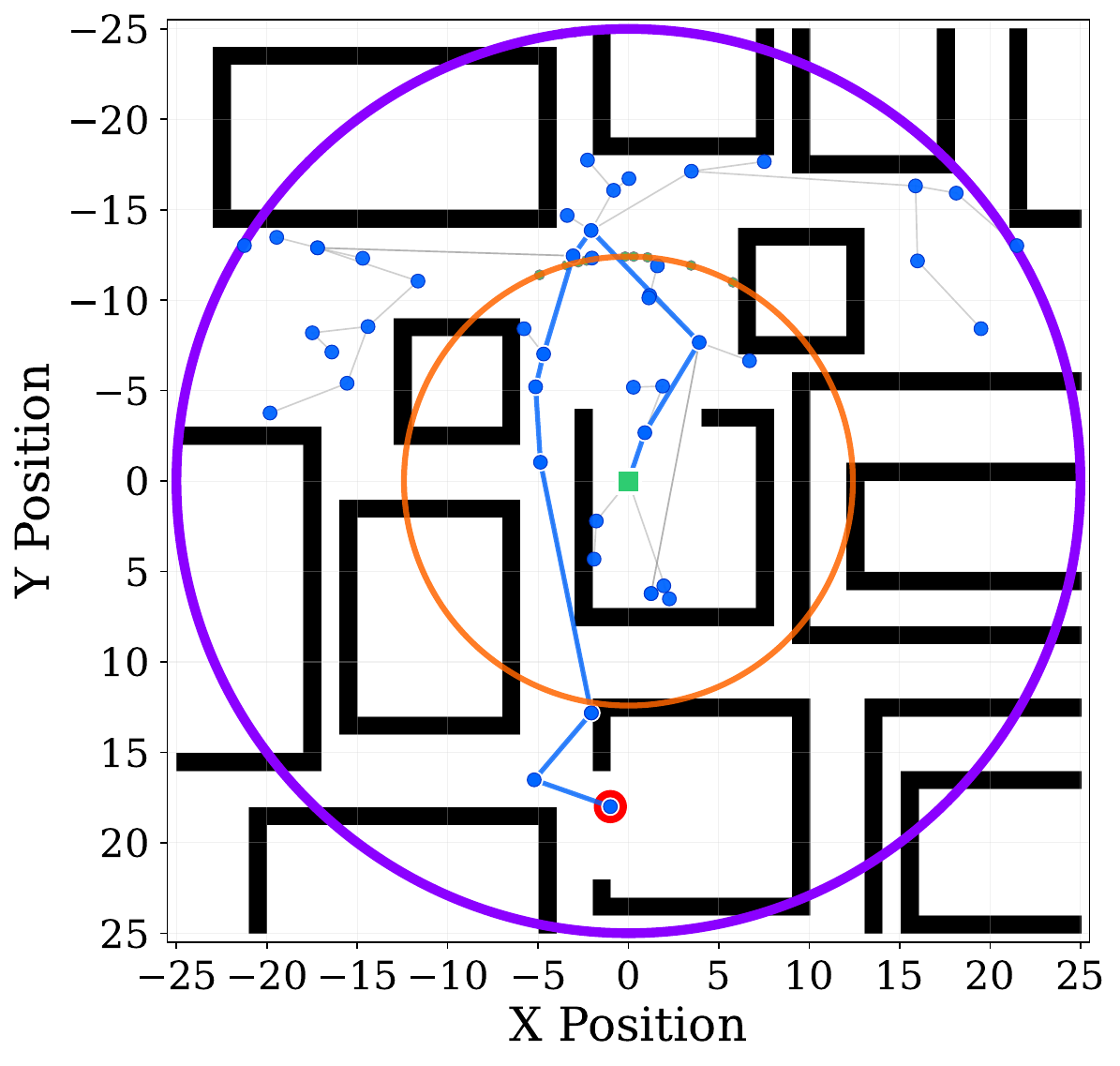} \par
    \end{minipage}
    \begin{minipage}[t]{0.33\linewidth}
        \centering
        \includegraphics[width=0.90\linewidth]{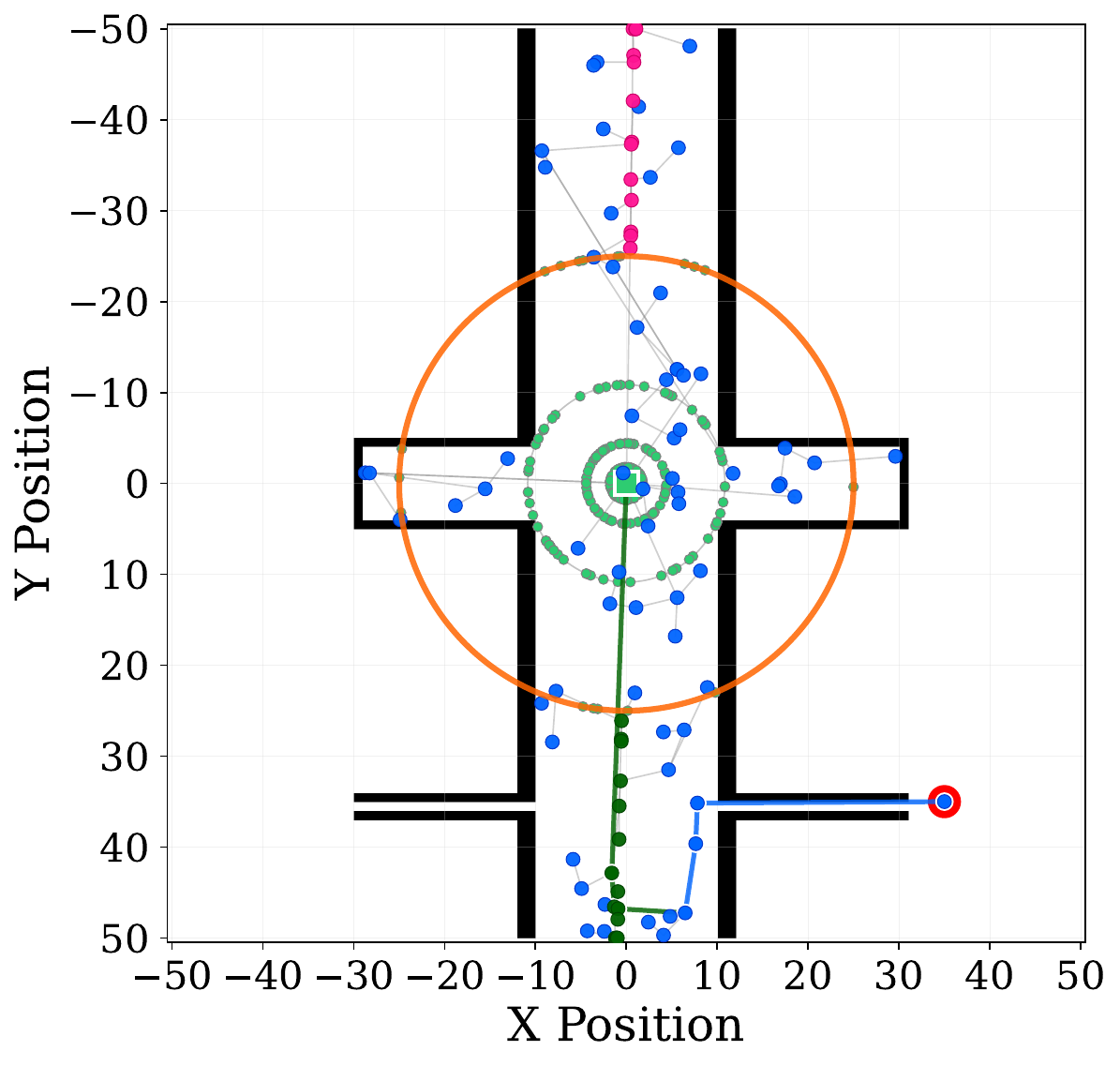}
        \includegraphics[width=0.90\linewidth]{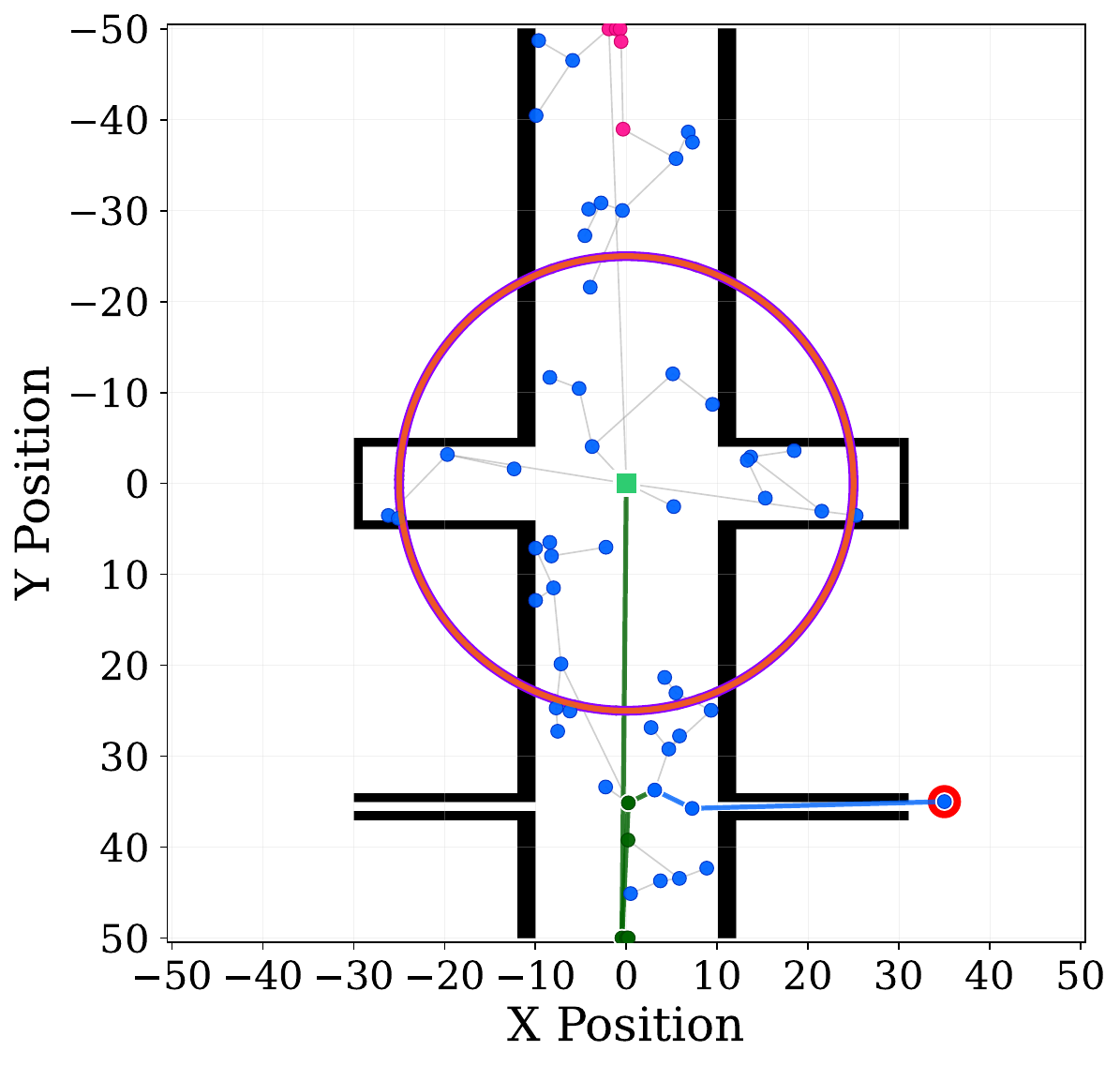}
    \end{minipage}
    \caption{Final sampling results across environments (radius set to 1e-6 vs. 25.0).
Final radius from grow-shrink shown as \legendline{orange}, initial radius as \legendline{violet}.
  Tree edges shown as \legendline{gray}.
  Valid samples: \cbox{blue} uniform, \cbox{green!60!black} positive principal component, \cbox{magenta} negative principal component, \cbox{green!70} scale-invariant sampler.
  Start marked with \csquare{green!70}, goal with \cmark{red}.
}
    \label{fig:env-final-grid}
\end{figure}
To show the results, we plot the success rate from zero to one hundred percent over time in log scale, as shown in Fig.~\ref{fig:results}. It can be seen that MAB-RRT solves all eight scenarios with $100$\% success rate, meaning that MAB-RRT found a solution in every single run. In the Socket environment, MAB-RRT reaches $100$ percent with over one order of magnitude (OoM) better runtime compared to the next best planners (RRT + Obstacle Sampling and RRT + Bridge Sampling). In the Eye Bolt scenario, MAB-RRT and BFS reach $100$ percent success rate, with MAB-RRT outperforming BFS by over 1 OoM. In the U-Bolt environment, only MAB-RRT reaches $100$ percent, while BFS only reaches $30$ percent success rate until the timeout. In Motor Flange, the situation is similar with MAB-RRT reaching $100$ percent and BFS reaching $30$ percent. In Gear Reducer, only MAB-RRT reaches $100$ percent, while planners RRT + Gaussian Sampling reaches $50$ percent, RRT + Bridge Sampling reaches $40$ percent, and RRT + Obstacle Sampling reaches $20$ percent. In the T-Bolt and Ring Bolt scenarios, MAB-RRT is the only planner reaching $100$ percent, each time with a runtime below $10^1$, which is 1 OoM below timeout. Finally, in Cross-Pin Connector, MAB-RRT, MateVec-TRRT, and BK-RRT reach $100$, while BFS and RRT + Obstacle Sampling reach $70$, RRT + Gaussian Sampling reaches $40$, and RRT + Bridge Sampling reaches $30$ percent. Again MAB-RRT outperforms the next best planner (BK-RRT) by 1 OoM.
In terms of runtime, MAB-RRT outperforms the next best planner or the timeout by at least 1 OoM on seven out of eight scenarios, with the exception of the U-Bolt scenario.
%%%%%%%%%%%%%%%%%%%%%%%%%%%%%%%%%%%%%%%%%%%%%%%%%%%%%%%%%%%%%
\subsection{Multi-Arm Bandit Dynamics}

\begin{figure}[!ht]
\centering

\begin{subfigure}[t]{0.48\linewidth}
    \centering
    \includegraphics[width=\linewidth]{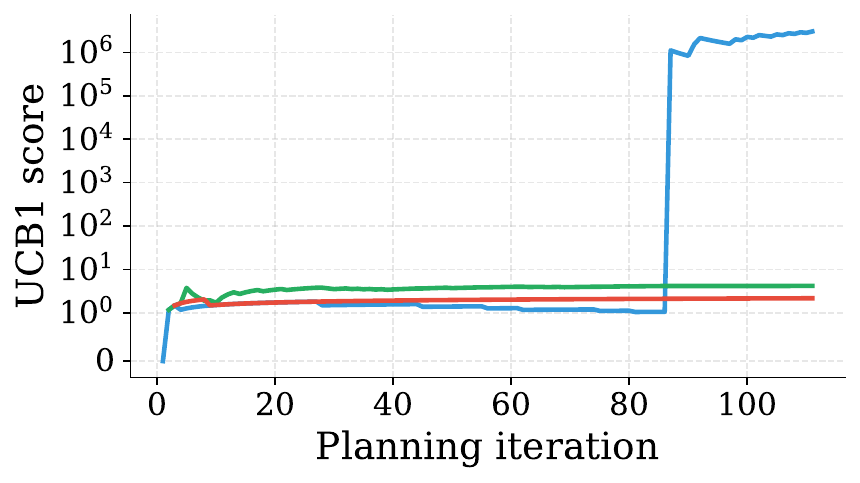}
    \caption{UCB scores (symlog scale).}
    \label{fig:mab_ucb}
\end{subfigure}
\hfill
\begin{subfigure}[t]{0.48\linewidth}
    \centering
    \includegraphics[width=\linewidth]{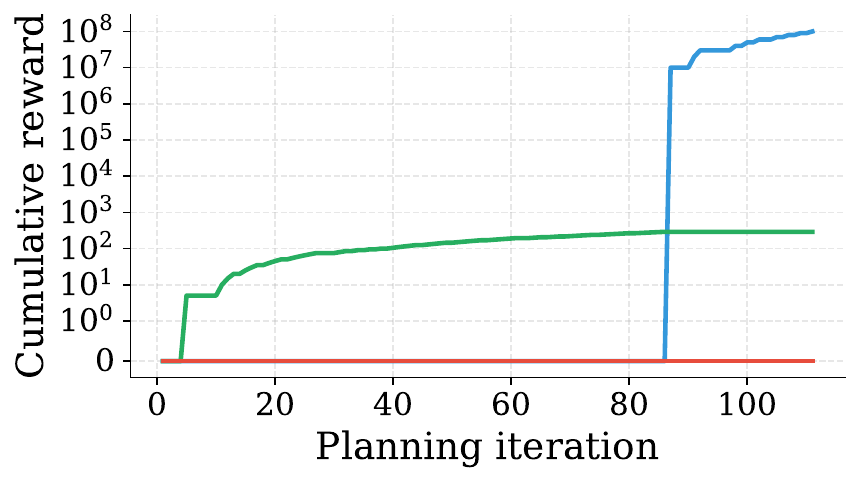}
    \caption{Cumulative rewards (symlog scale).}
    \label{fig:mab_rewards}
\end{subfigure}

\caption{T-Bolt Experiment: MAB dynamics during planning. (a)~UCB scores adjust dynamically as the planner gathers reward information. (b)~Rapid reward accumulation for cylinder arms indicates their effectiveness in generating valid samples within the narrow passage, while the uniform arm's reward surges once free space is reached. \cbox{blue}~Uniform, \cbox{green}~PCA positive, \cbox{red}~PCA negative.}
\label{fig:mab_dynamics}
\end{figure}

To verify that the multi-arm bandit correctly changes between arms over time, we run it on the T-Bolt experiment from Fig.~\ref{fig:env-comparison}. We plot both the UCB scores over time and the cumulative rewards per arm. This is depicted in Fig.~\ref{fig:mab_dynamics}. We can see that the cumulative rewards stays zero for uniform sampling, capturing the fact that uniform samples lead to direct collision. The reward for PCA positive is growing over time which reflects the fact that the robot can escape into only one direction from the tunnel. Once the robot has escaped (around iteration 85), there is a sharp uptick in reward for the uniform sampler, which reflects the growing of the tree into the open, free space.

The UCB scores closely follow this trend. In the beginning, there is a sharp decrease in UCB score for uniform sampling, reflecting the invalidity of the samples. Over time, the UCB scores for uniform and PCA negative tend towards similar values (${\sim}1.0$–2.0), so that uniform samples are still occasionally tried to verify that the tree has not yet reached the open space. Finally, at iteration 85, the UCB score has a sharp increase reflecting the switch to the uniform sampling scheme.

%%%%%%%%%%%%%%%%%%%%%%%%%%%%%%%%%%%%%%%%%%%%%%%%%%%%%%%%%%%%%
\subsection{Robustness of Scale-Invariant Sampling}

Another important aspect of our framework is the grow-shrink algorithm to find a useful high-information entropy scale. In this section, we like to show that this algorithm is robust against different initial radii. For this, we created three environments as depicted in Fig.~\ref{fig:env-final-grid}. Each row shows one environment, whereby the left image shows a starting radius of \num{1e-6} and the right image shows an initial radius of 25. For each scenario, we showcase the initial radius (violet circle) and the final radius after applying grow-shrink (orange circle). It can be seen that both initial radii faithfully converge to a similar radius at which roughly half of the samples are reachable. We also showcase the resulting trees with start (light green) and goal (red), where we show valid samples from uniform sampling (blue), samples from the principal component sampler with positive (green) and negative (magenta) direction, and initial samples from grow-shrink (light green). 

%\EnvGridFigure
%{burnin}
%{Burn-in spatial sampling comparison across environments (radius set to 1e-6 vs. 25.0).
%Concentric circles show tested radii during adaptive search, with final accepted radius as \legendline{green!70!black}.
%  Samples: \cbox{green!70} valid (motion from origin collision-free), \cbox{red} invalid (motion from origin blocked).
%  Origin marked with black dot.
%}
%{fig:env-burnin-grid}

%\EnvGridFigure
%{burnin_convergence_graph}
%{Burn-in convergence comparison across environments (radius set to 1e-6 vs. 25.0).
% \legendline{blue} validity rate (\%, left axis), \legendline{red} radius value (right axis).
%  Target range bounded by \legendline{red!70!black} min threshold and \legendline{green!60!black} max threshold.
%  Shaded region indicates acceptable validity range.}
%{fig:env-convergence-grid}

\section{Discussion and Conclusion}

We presented scale-invariant sampling, a novel sampling strategy for object extraction from tight, narrow passages. Our sampling strategy is based upon two methods. First, we explore the local space around an object's start position to find a scale where the information entropy is largest. This scale is maximally useful to find valuable samples. By using a grow-shrink method, we can robustly and quickly find such a scale. Second, we exploit the computed scale by leveraging the principal direction in the data, thereby finding a sampling bias which is helpful to robustly extract objects by finding the right directions inside a narrow passage. Finally, the samplers are integrated into a multi-arm bandit RRT (MAB-RRT)~\cite{faroni2023motion} which switches between different sampling strategies depending on the rewards obtained. 

The results show that scale-invariant MAB-RRT can successfully solve complex disassembly tasks. This involves sockets, gears, multiple bolt types, and connector pins. In our experiments, scale-invariant MAB-RRT outperforms similar biased sampling approaches and is the only planner which can reach a 100\% success rate on all scenarios. 

While the results are promising, there are still some open questions. First, while PCA could be used for arbitrary spaces and objects, we have not yet demonstrated this. It would be important to verify that this works also on challenging scenarios where objects have to be rotated (e.g. screws, keys), where longer narrow passages are present (e.g. pipes, rods, axles), where objects are deformable (e.g. pulling out cables), or where narrow passages are non-linear (e.g. buzz wire game~\cite{dorussen2021learning}). Second, we believe it would be important to analyze the convergence properties of the high information entropy scale search. While the planners work robustly, there is currently no guarantee that it will converge in every situation. 
Third, it is important to integrate our algorithm into a larger system for disassembly tasks~\cite{tian2025fabrica}, so that we can show that this can be used in combination with a real robot for complex object extraction tasks.

Despite limitations, scale-invariant MAB-RRT is a probabilistically complete and efficient planner which works reliably for complex object extraction tasks. We believe it is therefore an important component of a larger task and motion planning system~\cite{Bayraktar2023RAL} for automatic object disassembly.

% ---------- Bibliography ----------
\bibliographystyle{splncs04}
\bibliography{bib/references}

\end{document}